\newif \ifShowEmail
\ShowEmailtrue
\newif \ifShowORCID
\ShowORCIDfalse
\documentclass{article}
\usepackage[final]{acl}
\usepackage{times}
\usepackage{latexsym}
\usepackage[T1]{fontenc}
\usepackage[utf8]{inputenc}
\usepackage[htt]{hyphenat}
\usepackage{graphicx}
\usepackage{tabularx}
\usepackage{microtype}
\usepackage{makecell}
\usepackage{multirow}
\usepackage{interval}
\intervalconfig{ soft open fences, }
\usepackage[cache=false]{minted}
\usepackage[autolanguage]{numprint}
\usepackage[normalem]{ulem}
\usepackage{diagbox}
\usepackage{syntax}
\usepackage{tcolorbox}
\tcbuselibrary{skins,breakable}
\usepackage{pifont}
\usepackage{hyperref}
\usepackage{amsmath}
\usepackage{amssymb}
\usepackage{booktabs}
\usepackage{wrapfig}
\usepackage{subcaption}
\usepackage{listings}
\usepackage{enumitem}
\usepackage{pgf}
\setlist[itemize]{leftmargin=15pt}

\newcommand{\para}[1]{\smallskip\noindent {\bf #1} }

\begin{document}
\newif \ifBlinded
\Blindedtrue

\setminted[json]{
mathescape,
linenos,
numbersep=-4pt,
gobble=2,
frame=lines,
framesep=1mm,
breaklines,
fontsize=\footnotesize,
escapeinside=||
}

\setminted[shell]{
mathescape,
linenos,
numbersep=-4pt,
gobble=2,
frame=lines,
framesep=1mm,
breaklines,
fontsize=\footnotesize,
escapeinside=||
}

\setminted[python]{
mathescape,
linenos,
numbersep=-4pt,
gobble=2,
frame=lines,
framesep=1mm,
breaklines,
fontsize=\footnotesize,
escapeinside=||
}

\setminted[c++]{
mathescape,
linenos,
numbersep=-4pt,
gobble=2,
frame=lines,
framesep=1mm,
breaklines,
fontsize=\footnotesize,
escapeinside=||
}

\newcommand{\toolname}{\textsc{Nsvif}}
\newcommand{\gepabaseline}{\textsc{GEPA-CoT}}
\newcommand{\cotbaseline}{\textsc{Conv-CoT}}
\newcommand{\toolnamesolver}{\textsc{Nsvif-Neu}}
\newcommand{\bench}{\textsc{VifBench}}
\newcommand{\baseline}{LLM-as-a-judge}
\newcommand{\autoform}{Autoformalization}
\newcommand{\DL}{\textit{DL}}

\newcommand{\red}{\color{red}}
\newcommand{\mgt}{\color{magenta}}

\newcommand{\fscore}[1]{%
    \pgfmathparse{int(100 * 1.05^(#1 - 100))}
    \edef\temp{\noexpand\colorbox{YellowGreen!\pgfmathresult}}%
    \temp{#1\%}
}

\newcommand{\symbolimg}[2][0.3cm]{%
  \ensuremath{\vcenter{\hbox{\includegraphics[height=#1]{#2}}}}%
}

\newcommand{\openai}{\symbolimg[0.35cm]{icons/openai.png}}
\newcommand{\deepseek}{\symbolimg[0.35cm]{icons/deepseek.png}}
\newcommand{\qwen}{\symbolimg[0.35cm]{icons/qwen.png}}

\newtcolorbox{nsviupromptparser}[1]{
    colframe=lime!40!black,colbacktitle=lime!20,
    fontupper=\ttfamily\tiny,
    coltitle=black,boxrule=1.2pt,enhanced,arc=2pt,fonttitle=\bfseries\scriptsize,
    title={\small \toolname{} Prompt \hfill {#1}},
    breakable
}

\newtcolorbox{nsviupromptexecutor}[1]{
    colframe=cyan!40!black,colbacktitle=cyan!20,
    fontupper=\ttfamily\tiny,
    coltitle=black,boxrule=1.2pt,enhanced,arc=2pt,fonttitle=\bfseries\scriptsize,
    title={\small \toolname{} Prompt \hfill {#1}},
    breakable
}

\newtcolorbox{nsviupromptsolver}[1]{
    colframe=purple!40!black,colbacktitle=purple!20,
    fontupper=\ttfamily\tiny,
    coltitle=black,boxrule=1.2pt,enhanced,arc=2pt,fonttitle=\bfseries\scriptsize,
    title={\small \toolname{} Prompt \hfill {#1}},
    breakable
}

\newtcolorbox{baselineprompt}[1]{
    colframe=orange!40!black,colbacktitle=orange!20,
    fontupper=\ttfamily\tiny,
    coltitle=black,boxrule=1.2pt,enhanced,arc=2pt,fonttitle=\bfseries\scriptsize,
    title={\small Baseline LLM-as-a-Verifier Prompt \hfill {#1}},
    breakable
}

\newtcolorbox{vifbenchprompt}[1]{
    colframe=yellow!80!black,
    fontupper=\ttfamily\tiny,
    coltitle=black,boxrule=1.2pt,enhanced,arc=2pt,fonttitle=\bfseries\scriptsize,
    title={\small VIFBench Prompt \hfill {#1}},
    breakable
}

\newtcolorbox{gepaexample}[1]{
    colframe=yellow!80!black,
    fontupper=\ttfamily\tiny,
    coltitle=black,boxrule=1.2pt,enhanced,arc=2pt,fonttitle=\bfseries\scriptsize,
    title={\small {#1} \hfill },
    breakable
}

\newtcolorbox{nsvifsolveronlyexample}[1]{
    colframe=yellow!80!black,
    fontupper=\ttfamily\tiny,
    coltitle=black,boxrule=1.2pt,enhanced,arc=2pt,fonttitle=\bfseries\scriptsize,
    title={\small {#1} \hfill },
    breakable
}

\newtcolorbox{travelagent}[1]{
    colframe=cyan!80!black,
    fontupper=\ttfamily\tiny,
    coltitle=black,boxrule=1.2pt,enhanced,arc=2pt,fonttitle=\bfseries\scriptsize,
    title={\small {#1} \hfill },
    breakable
}

\definecolor{codegreen}{rgb}{0,0.6,0}
\definecolor{codegray}{rgb}{0.5,0.5,0.5}
\definecolor{codepurple}{rgb}{0.58,0,0.82}
\definecolor{backcolour}{rgb}{0.95,0.95,0.92}
\definecolor{lightgreen}{rgb}{0.66, 0.84, 0.61}

\lstdefinestyle{textstyle}{
    backgroundcolor=\color{backcolour},   
    commentstyle=\color{codegreen},
    keywordstyle=\color{black},
    numberstyle=\tiny\color{black},
    stringstyle=\color{black},
    basicstyle=\ttfamily\footnotesize,
    breakatwhitespace=false,         
    breaklines=true,                 
    captionpos=b,                    
    keepspaces=true,                 
    numbers=left,                    
    numbersep=5pt,                  
    showspaces=false,                
    showstringspaces=false,
    showtabs=false,                  
    tabsize=2,
}

\lstdefinestyle{codestyle}{
    backgroundcolor=\color{backcolour},   
    commentstyle=\color{codegreen},
    keywordstyle=\color{magenta},
    numberstyle=\tiny\color{codegray},
    stringstyle=\color{codepurple},
    basicstyle=\ttfamily\footnotesize,
    breakatwhitespace=false,         
    breaklines=true,                 
    captionpos=b,                    
    keepspaces=true,                 
    numbers=left,                    
    numbersep=5pt,                  
    showspaces=false,                
    showstringspaces=false,
    showtabs=false,                  
    tabsize=2
}

\title{\Large Neuro-Symbolic Verification on Instruction Following of LLMs}

\author{
	Yiming Su$^{1}$\thanks{Equal Contribution. This paper is a result of an open-source project. Kunzhao Xu is a PhD student. This work was done during Kunzhao's internship at Microsoft Research.},
	Kunzhao Xu$^{2}$\footnotemark[1],
	Yanjie Gao$^{3}$\thanks{Corresponding author: Yanjie Gao \\ (E-mail: yanjga@microsoft.com)},
	\textbf{Fan Yang}${^3}$, \\
	\textbf{Cheng Li}${^2}$,
	\textbf{Mao Yang}${^3}$,
	\textbf{Tianyin Xu}$^{1}$ \\
	$^{1}$University of Illinois Urbana-Champaign\\
	$^{2}$University of Science and Technology of China\\
	$^{3}$Microsoft Research
}
	
\maketitle
\begin{abstract}

A fundamental problem of applying Large Language Models (LLMs) to important applications is that LLMs do not always follow instructions, and violations are often hard to observe or check.
In LLM-based agentic workflows, such violations can propagate and amplify along reasoning chains, causing task failures and system incidents.
This paper presents \toolname{}, a neuro-symbolic framework for verifying whether an LLM's output follows instructions used to prompt the LLM.
\toolname{} is a universal, general-purpose verifier; it makes no assumption on the instruction or the LLM.
\toolname{} formulates instruction-following verification as a constraint-satisfaction problem by modeling user instructions as {\it constraints}.
\toolname{} models both logic and semantic constraints; constraint-solving is done by a unified solver that orchestrates logic reasoning and semantic analysis.
To evaluate \toolname{}, we develop \bench{}, a new benchmark for instruction-following verifiers with fine-grained data labels.
Experiments show that \toolname{} significantly outperforms LLM-based approaches and provides interpretable feedback.
We also show that feedback of \toolname{} helps improve LLMs' instruction following capability without post-training.

\end{abstract}

\section{Introduction}
\label{sec:intro}

Large Language Models (LLMs) are increasingly used in real-world applications. However, a fundamental problem of applying LLMs to important applications is that LLMs do not always follow user instructions---LLMs are reported to constantly misunderstand or ignore critical parts of user instructions~\citep{jaroslawicz2025instructionsllmsfollowonce,ifsurvey}. Such violations can lead to unsafe decisions, incorrect outputs, and system failures. In agentic use cases, the impact of violations can propagate and amplify along reasoning chains and workflows~\citep{cemri2025multi,laban2025llmslostmultiturnconversation}.

Despite significant improvement of frontier LLMs, their ability of instruction following is still unreliable. Unfortunately, it is non-trivial to observe or check an LLM's instruction following behavior. The common approach, referred to as LLM-as-a-judge, uses an LLM to check whether an LLM's output follows instructions based on the input prompt~\citep{sirdeshmukh2025multichallengerealisticmultiturnconversation,reife,openai2025gpt5,geval,cemri2025multi}. However, as shown in \S\ref{sec:results}, existing LLM-based approaches are unreliable for verifying complex instructions. The difficulty in verifying instruction following behavior (1) makes it hard for agents to prevent violations from propagation~\citep{laban2025llmslostmultiturnconversation}, and (2) makes failure attribution hard~\citep{zhang2025agent}. 

\begin{table*}
  \centering
  \footnotesize
  \caption{Common types of violations of LLM instruction following.}
  \label{tab:error} 
  \vspace{-5pt}
  \def\arraystretch{1.2}%
    \begin{tabular}{|l|l|l|} 
      \hline
      \bf Category & \bf Type  & \bf Desc. \\
      \hline
      \hline
      \multirow{3}*{Logic}
      & Invariant violation &  Output violating invariants such as ordering or uniqueness rules described in the prompt \\
      \cline{2-3}
      & Structural violation &  Output not adhering to described schema or data type (e.g., JSON, XML, or template) \\
      \cline{2-3}
      & Invalid element & Output omitting required elements or including prohibited entities \\
      \cline{2-3}
      \hline
      \hline
      \multirow{2}*{Semantic}
      & Misinterpretation  & LLMs misunderstanding of instruction semantics and failures of resolving ambiguity \\
      \cline{2-3}
      & Pragmatic mismatch &  Output misaligning with the intent, stylistic requirements, or scope of the instruction \\
      \cline{2-3}
      \hline
    \end{tabular}%
\end{table*}

This paper presents {\toolname}, a neuro-symbolic framework for verifying instruction following of LLMs. \toolname{} aims to be universal and general-purpose---taking any output produced by an LLM, \toolname{} checks if the output follows the input instruction used to prompt the LLM. \toolname{} makes no assumption on the instruction or the LLM, and does not require additional information. We expected its usability to lead to wide use cases such as agentic workflows~\citep{qi2025agentifbenchmarkinginstructionfollowing}.

The key idea is to formalize instruction following verification as a {\it constraint-satisfaction problem} or {\it CSP}. Concretely, \toolname{} models instructions described in the prompt as a logic formula of {\it constraints} and checks whether the output of an LLM satisfies the formula. \toolname{} models two types of constraints: {\it logic} and {\it semantic} constraints. The former can be symbolically checked with logic reasoning (e.g., whether a paragraph contains a given keyword); the latter requires semantic analysis using neural approaches (e.g., whether a sentence has a positive tone). \toolname{} addresses the CSP by generating per-constraint checkers and solving them using a unified solver atop the Z3 SMT (Satisfiability Modulo Theories) solver~\citep{z3}.

\toolname{} has several advantages over state-of-the-art LLM-based approaches. First, it brings logic reasoning for instructions that can be and should be checked by traditional tools and symbolic analysis. Moreover, the decomposition of instructions into fine-grained constraints enables the divide-and-conquer strategy, which is more effective than directly verifying a composite instruction. Third, the CSP formulation makes the results significantly more interpretable---\toolname{} pinpoints the violated constraint and how the output fails it.

Note that semantic constraints are inherently subjective and it is hard for semantic analysis to draw absolute conclusions. For each individual semantic constraint, \toolname{} inherits the properties of the semantic analysis it uses. As a framework, \toolname{} can be equipped with any semantic analysis for semantic constraints. Following the common practice~\citep{reife}, our implementation uses LLMs to evaluate each individual semantic constraint. Other models with verified predictability~\citep{Albarghouthi2021} or specialized personalities can also be used. The key to \toolname{} is decomposing fine-grained constraints and applying semantic analysis and logic reasoning accordingly under a unified framework.

We develop \bench{}, a new benchmark that measures the effectiveness of instruction following verifiers like \toolname{}. \bench{} contains 820 labeled data, in the form of <instruction, output, result> tuples, where an instruction consists of a set of constraints and their relations, the output has detailed annotations on which constraints it violates (if any), the result asserts whether the output follows the instruction. Different from existing benchmarks~\cite{zeng2024evaluatinglargelanguagemodels,reife} that only provide one single result label, \bench{} provides fine-grained analysis on how a verifier reasons about instruction following behavior. We evaluate \toolname{} on \bench{}. Our results show that \toolname{} can significantly outperform widely used baselines that directly use LLMs as judges, demonstrating the effectiveness of the neuro-symbolic approach in verifying instruction following. We show that \toolname{} provides constraint-level feedback and improves LLM instruction following capability without post-training.

In summary, we make the following contributions:
\vspace{-3.5pt}
\begin{itemize}[itemsep=0.5pt]
	\item {\bf Approach.} A neuro-symbolic approach for instruction following verification,
	which formulates verification as a constraint-satisfaction problem and
	model both logic and semantic constraints.
	\item {\bf Tooling.} {\toolname}, a universal, general-purpose framework for
	verifying LLM outputs against instructions and improving
	LLM's instruction following capability without post-training.
	\item {\bf Benchmark.} {\bench}, a new benchmark and data synthesis toolchain
	for evaluating verification techniques on instruction following of LLMs.
\end{itemize}

\section{Background}
\label{sec:bg}

\subsection{Instruction Following of LLMs}
\label{sec:bg:if}

As a basic ability, instruction following is unreliable even in today's frontier LLMs~\cite{openai2025gpt5}. Prior studies~\cite{ifsurvey,jaroslawicz2025instructionsllmsfollowonce,laban2025llmslostmultiturnconversation,sirdeshmukh2025multichallengerealisticmultiturnconversation} reported various kinds of LLM behavior that fail to follow instructions; Table~\ref{tab:error} summarizes a few common types. Therefore, effective verification of instruction following is desired.

Despite significant efforts on benchmarking instruction following ability of LLMs~\citep{he2024largelanguagemodelsunderstand,wu2025lifbenchevaluatinginstructionfollowing,qi2025agentifbenchmarkinginstructionfollowing,if256,IFEval,EIFBENCH,complexbench,ifsurvey,qin2024infobenchevaluatinginstructionfollowing}, we are in search of a general-purpose technique that can effectively, reliably verify whether an output of an LLM follows the input instruction. We find that evaluation techniques used by existing instruction following benchmarks either use domain-specific oracles~\cite{laban2025llmslostmultiturnconversation,jaroslawicz2025instructionsllmsfollowonce,complexbench} or use LLMs as evaluators~\cite{zeng2024evaluatinglargelanguagemodels,reife}. Domain-specific oracles often assume the types of instructions used in the benchmarks and thus are not applicable to cases not covered in the benchmark. Existing LLM-based evaluators, on the other hand, is unreliable, as shown in our evaluation (\S\ref{sec:results}).

Note that \toolname{} focuses on verification of instruction following, rather than evaluation of natural language generation~\cite{gao-etal-2025-llm}. In other words, \toolname{} checks behavior correctness of LLMs with respect to constraints and intent, not linguistic quality per se.

\subsection{Constrained Decoding}
\label{sec:constrained-decoding}

\toolname{} applies to instructions in both natural language prompts and structured inputs written in declarative languages or schemas for constrained decoding. For the former, \toolname{} extracts constraints from input prompts. For the latter, DSPy~\cite{dspy} and PDL~\cite{pdl} allows users to specify type constraints, including primitive data types and natural-language typed signatures; LMQL~\cite{lmql} provides built-in functions that allow users to specify logical constraints such as word count or last token in a sequence. \toolname{} can directly interface with these tools; it verifies type and structure constraints by synthesizing type and format checkers. \toolname{} goes beyond type and structure constraints. Mellea~\cite{mellea} which allows users to define requirements of the LLM output. Mellea uses LLMs as verifiers unless user-defined functions are provided. Different from Mellea, \toolname{} does not need to explicitly define constraints {\it a priori}. Moreover, \toolname{} does not directly apply LLMs to verify all constraints and their compositions.

\subsection{Formal Reasoning of LLMs and Agents}

\toolname{} applies formal reasoning to verify and reshape an LLM's outputs based on user instructions. The principle echos recent practices of specifying model and agent behavior~\citep{ibm-formal-agent-spec,formal-llm}. Differently, \toolname{} focuses on instruction following behavior of LLMs, instead of other behavior like tool calling and multi-agent coordination. \toolname{} can be used as a step-wise verifier in an agentic workflow, where each agent step and its output can be checked by \toolname{}.

\toolname{} uses LLMs as helpers to formulate the verification problem into first-order logic predicates and solve it using a SMT solver. The principle is applied in related work like LINC~\citep{linc} and Logic-FM~\citep{logic-fm} which use LLMs to translate natural language premises and conclusions into first-order logic statements and use a theorem prover to prove the statements. Differently, \toolname{} needs to formulate the premises in the verification process as it needs to individually check whether each constraint is satisfied.
\section{\toolname{}: A Neuro-Symbolic Framework}
\label{sec:method}

\toolname{} is a universal, general-purpose framework for verifying whether an output $O$ produced by an LLM follows the instruction $I$ used to prompt the LLM. It takes any arbitrary $\langle I, O \rangle$ and returns a boolean result. It makes no assumption on $I$, $O$, or the LLM, and does not need any hints or other information.

\toolname{} is built atop {\it neuro-symbolic} principles. First, inspired by prior work on modeling an instruction as a composition of constraints~\cite{complexbench,ifbench,jiang-etal-2024-followbench,IFEval,pham-etal-2024-suri,yuan-etal-2025-following}, \toolname{} models an instruction as a formula in first-order logic where the predicates are constraints. In essence, \toolname{} formulates the verification problem as a {\it constraint-satisfaction problem} or {\it CSP}. In this way, \toolname{} not only reasons about constraints individually but also their compositions collectively. The use of first-order logic makes \toolname{} sufficiently expressive. It encodes various kinds of constraints into a {\it unified} formal reasoning framework and enabling reasoning tools like SMT (Satisfiability Modulo Theories) solvers (our implementation uses Z3~\cite{z3}) to rigorously check if an given output $O$ follows the instruction.

The formulation enables specialized reasoning of individual constraints based on their inherent properties. Real-world constraints are different. Some are about {\it logic}, which specify lexical, format, and structure requirements, e.g., whether a paragraph contains a keyword. Such constraints should be verified by logical reasoning. The others are {\it semantic}, which specify requirements on topic, tone, style, sentiment, e.g., whether a paragraph is on the topic of science. Such constraints should be verified by semantic analysis. \toolname{}'s divide-and-conquer strategy makes it significantly more effective than verifying the entirety of the instruction.

Concretely, for a given instruction, \toolname{} first formulates the CSP by using an LLM to translate the instruction into a first-order logic formula (\S\ref{sec:formulate}). It then checks every individual constraint against the output. For each constraint, \toolname{} first uses an LLM to classify whether the constraint is a logic constraint or a semantic one. For each logic constraint, \toolname{} uses an LLM to generate code that check the constraint against the output. For semantic analysis, \toolname{} prompts an LLM as the judge. Lastly, \toolname{} solves the CSP as a Z3 program and output the final result.

\begin{figure*}[h]
    \centering
    \includegraphics[width=\textwidth]{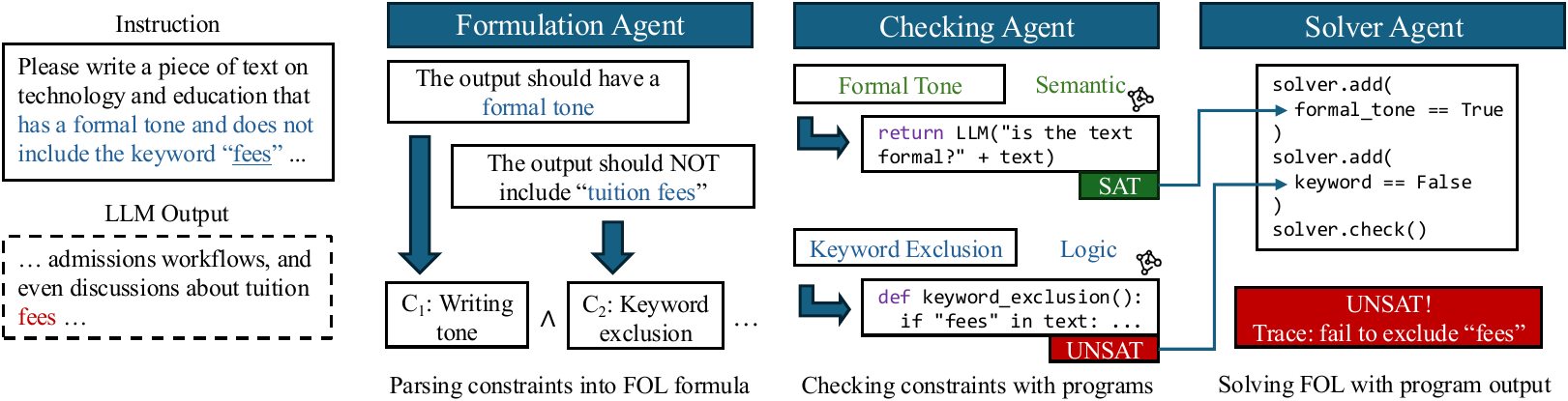}
    \vspace{-20pt}
    \caption{Overview of the \toolname{} framework and the verification workflow.}
    \vspace{-10pt}
    \label{fig:nsvif-overview}
\end{figure*}

As a neuro-symbolic approach, \toolname{} uses LLMs for a series of tasks, including (1) extracting constraints from the instruction and formulating the instruction into a first-order logic formula with constraints as predicates~\citep{linc,specgen}, (2) classifying constraints into logic or semantic types~\citep{carp}, (3) for logic constraints, generating the checker code~\citep{linc, position-fm,jimenez2024swebench, swebench-verified}, and (4) for semantic constraints, checking if the output satisfies the constraints. Therefore, like any LLM-based approaches, \toolname{} also subjects to hallucination behaviors of LLMs. On the other hand, by breaking down instruction following verification into specific, well-defined tasks, \toolname{} significantly improves verification effectiveness and produces interpretable results.

\subsection{Problem Formulation}
\label{sec:formulate}

\toolname{} formulates the instruction following verification problem as a Constraint-Satisfaction Problem or CSP. For an input $\langle I, O \rangle$, \toolname{} verifies whether $O$ satisfies the constraints $C$ and their relations in a first-order logic formula $f(C)$ specified by $I$,
\[
\text{verify}(I, O) =
\begin{cases}
	\text{SAT} & \text{if } O \models f(C) \\
	\text{UNSAT} & \text{otherwise}
\end{cases}
\]
\toolname{} differentiates two types of constraints:

\begin{itemize}[itemsep=0.5pt]
	\item A {\it logic constraint}
	can be encoded using symbolic rules, regular expressions,
	or executable code. Satisfaction is checked via symbolic reasoning and/or constraint solving.
	The verification of logic constraints is objective based on first principles.
	
	\item A {\it semantic constraint} is hard to be verified by logic reasoning
	but requires semantic analysis using neural approaches.
	Although a semantic constraint lacks explicit structure,
	it implicitly encodes a learned boundary between satisfying and violating the constraint
	based on in-context examples or prompts.
	Note that the semantic analysis can be subjective, e.g., the same writing
	can be perceived differently by different people or LLMs.
	Therefore, the verification results are not absolute but with regard to the semantic analysis (e.g., the LLM used
	in \toolname{} and its inference parameters).
\end{itemize}

\toolname{} encodes all the constraints
into a unified first-order logic formula as a CSP using logical connectives
including NOT ($\neg$), AND ($\land$), and OR ($\lor$).
\toolname{} supports common connectives include negation, disjunction, conjunction, implication,
and equivalence, which can express all the specific composition types
in prior work~\citep{complexbench,jaroslawicz2025instructionsllmsfollowonce,IFEval,sirdeshmukh2025multichallengerealisticmultiturnconversation}.
The logic formula is then checked by an SMT solver.

\subsection{Implementation}
\label{sec:impl}

As shown in Figure~\ref{fig:nsvif-overview}, we implement \toolname{} as a multi-agent system that orchestrates a formulation agent, a checking agent, and a solver agent. \toolname{} takes a <$I$, $O$> pair as input and outputs a boolean result which tells whether $O$ follows $I$, and if not, a detailed explanation.

\para{Formulation agent.}
The agent prompts an LLM to formulate the CSP based on the input instruction. The prompt asks the LLM to identify constraints in the instruction and analyze their relations. For each identified constraint, the agent gives a one-sentence summary of the constraint. The prompt then instructs the LLM to analyze the relations of the constraints, asking for a first-order logic formula of the constraints and encoding it into a Z3 program. We include several in-context learning examples on parsing constraints from example texts in the prompt to demonstrate instructions with both logic and semantic constraints with their desired outputs. We also include a chain-of-thought reasoning instruction to help the agent reason through the parsing process. The formulation agent produces two items: (1) a list of constraint summaries as text, each corresponds to a constraint in the instruction and (2) a Z3 program that encodes the first-order logic formula of constraints. If the constraints and their relations are already specified using constrained decoding frameworks (\S\ref{sec:constrained-decoding}), the agent only needs to translate them into a Z3 program.

\para{Checking agent.}
For each constraint summary received, the agent generates code to check whether the output $O$ satisfies the constraint. The agent first prompts an LLM to classify whether a constraint is a logic or a semantic constraint. In this step, we input the agent with the original <$I$, $O$> pair, the constraint summary, and the Z3 program. For each constraint, the agent first generates a checking plan: (1) the constraint summary and its type, (2) the input and output of the checker code, and (3) the expected behavior of the checker. The agent then uses another prompt to generate the checker code based on the plan. For each logic constraint, the agent generates an executable Python program to check whether the output satisfies the constraint. (Python is chosen because it is reported to be the best supported language by LLMs~\cite{yang2025swesmith}). For each semantic constraint, the agent generates a short, zero-shot LLM prompt tailored to the constraint to check whether the output satisfies the constraint. The checking agent then executes all the programs and saves the results into files. Note that the generated checker code can fail (e.g., throwing runtime exceptions); we allow multiple self-reflections (three by default) for the LLM to fix or improve the code. If the LLM fails to fix errors within given budget, it gives up and falls back to use LLMs directly to evaluate the constraint, which essentially treats the logic constraint as a semantic constraint. We include in-context learning examples that showcase example logic and semantic constraints. We also require chain-of-thought reasoning on how a constraint should be verified to ground the agent in the program generation task. The prompt for generating checker code can be found in Appendix~\ref{sec:prompt}.

The agent aggregates all individual constraint results and execute the Z3 program to produce the final verification result. The agent produces both a boolean result (\texttt{sat} or \texttt{unsat}) and an \textit{explanation} for interpretability. The explanation provides detailed information that pinpoints the violated constraints and the logic formula that leads to the final result. For a violated logic constraint, it includes the checker code that invalidates $O$. For a violated semantic constraint, it includes the LLM's response and reasoning traces.

\para{Extensibility.}
The modularity of \toolname{} makes it highly extendable. First, we allow users to customize LLMs across agents and within one agent, as different LLMs have specialized capabilities. For example, the checking agent can benefit from strong coding LLMs especially if the checker code is large and complex. The LLMs for evaluating semantic constraints can be customized based on concrete domains and use cases.

The techniques for checking logic and semantic constraints can be extended independently, with the emerging of new instruction types or domains. Specifically, \toolname{} exposes an interface that allows users to add custom constraints and corresponding checkers to help incorporate domain-specific knowledge and context-sensitive analysis. To do so, users need to package the checkers as an executable function in Python---the checking agent can discover custom functions and use them in corresponding checkers.

We also plan to equip \toolname{} with advanced tools (e.g., for querying system states) for use cases that require external information or advanced utilities (see \S\ref{sec:conclusion}).

\subsection{Use Cases}

\toolname{} targets LLM use cases that require verification of instruction following behavior. One target use case is agentic workflows, where agents frequently instruct LLMs and do so iteratively. Prior work~\cite{cemri2025multi} shows that instruction following is critical in agentic workflows as violations can propagate and amplify along reasoning chains and workflows. \toolname{} can serve as a per-step verifier to ensure every output from the LLM follows the instruction issued by the agent. Upon violations, \toolname{}'s interpretable explanations are useful feedback as explicit learning signals for LLMs to better internalize complex constraints and correct violations (\S\ref{sec:eval:agent}). Such a verification-feedback-improvement loop could enhance the LLM's instruction-following ability.

\section{{\bench{}}}
\label{sec:bench}

\bench{} is a new benchmark that evaluates the effectiveness of verification techniques of LLM instruction following. It contains 820 data entries in the form of <instruction, output, result> tuples, where the instruction contains multiple constraints with fine-grained labels and the result is either \texttt{sat} or \texttt{unsat}. A verifier under evaluation (\toolname{} or LLM judges) is asked to verify whether the output satisfies the instruction; the answer is checked against the result label.

We decided to create a new benchmark, because existing benchmarks for instruction following verification, namely LLMBar~\cite{zeng2024evaluatinglargelanguagemodels} and ReIFE~\cite{reife} (which uses LLMBar) are limited. LLMBar formulates verification of instruction following as a {\it classification problem}. Each instruction is paired with two outputs---one is preferred over the other as per the instruction. An LLM-as-a-judge is prompted to choose between the two. We attempted to use LLMBar in the evaluation of \toolname{}, but found it unreliable---preference does not indicate instruction following. A preferred output may still violate constraints and less preferred ones can satisfy all constraints. LLMBar does not provide fine-grained information that explains why and how an output violates the instruction.

\begin{figure}
\centering
\includegraphics[width=\columnwidth]{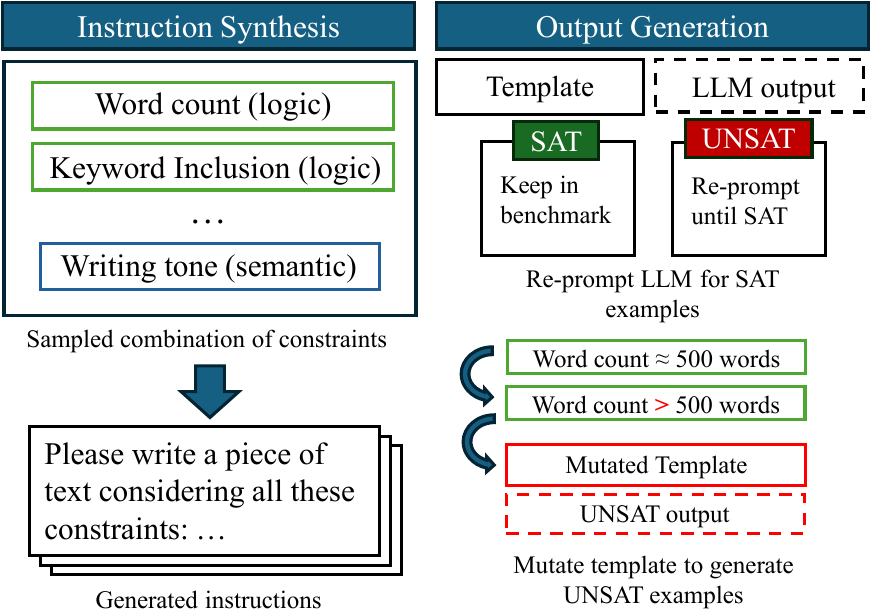}
\vspace{-17.5pt}
\caption{Overview of \bench{}'s data generation}
\label{fig:benchmark-datapoint-example}
\end{figure}

\bench{} addresses these limitations by (1) synthesizing instruction from well-defined constraints, and (2) providing fine-grained information whether the output satisfies each constraint and the reasons of violations if any. \bench{} includes both \texttt{sat} and \texttt{unsat} outputs that cover common categories in Table~\ref{tab:error}. Figure~\ref{fig:benchmark-datapoint-example} shows the workflow of data generation of \bench{}.

\subsection{Instruction Synthesis}

The instruction synthesis follows the practice of ComplexBench~\cite{complexbench}. We define a set of constraints and compose them into complex instructions. \bench{} focuses on verifying instruction following and uses English writing tasks which do not require additional ability of LLMs (e.g., using external tools).

For constraints, we surveyed constraints from both instruction tuning datasets~\cite{oasst1, lmsys1m, wildchat} and instruction following benchmarks~\cite{complexbench,IFEval,qin2024infobenchevaluatinginstructionfollowing}. These constraints cover the ones mentioned in Table~\ref{tab:error}. For logic constraints, we include lexical and structural constraints such as word count and keyword inclusion. For semantic constraints, we included writing topic (as pragmatic mismatch) and writing tone (for misinterpretation). In total, \bench{} includes eight types of logic constraints and two types of semantic constraints, as shown in Table~\ref{tab:vifbench-const-table}.

Semantic constraints in \bench{} are described as desired attributes for the satisfying output text, e.g., The output text should be in a positive tone''. Logic constraints are further specified with specific values that attaches to specific writing topic'' values. We focus on writing tone and writing topic because we find that they are the least subjective constraints and thus help evaluate the correctness of the verification. Other semantic constraints such as clarity and fluency often solicit subjective responses, confounding verification results. (\bench{} focuses on validation of instruction following, not evaluation of natural language generation (see \S\ref{sec:bg:if}).)

We then synthesize instructions by composing constraints. \bench{} varies the number of constraints in one instruction, which represents the complexity of the instruction. For a chosen number of constraints in one instruction, we permute the logic and semantic constraints and their relations. Each instruction in \bench{} contains at least one semantic constraint and one logic constraint. By mixing both types of constraints, \bench{} evaluates whether a verifier can conduct both logic reasoning and semantic analysis. In summary, an instruction in \bench{} prompts LLMs to output a writing response with specific requirements.

\begin{table*}[ht]
\centering
\small
\setlength{\tabcolsep}{4.5pt}
\caption{Verification effectiveness of {\toolname} and the baseline on {\bench}}
\vspace{-5pt}
\label{tab:rq-vif}
\begin{tabular}{lcccccccc}
    \toprule
    \multirow{2}{*}{\textbf{Model}} & \multicolumn{4}{c}{\textbf{\toolname{}}} & \multicolumn{4}{c}{\bf Baseline (LLM-as-a-judge)} \\
    \cmidrule(lr){2-5} \cmidrule(lr){6-9}
     & \textbf{F1 score} & \textbf{Precision} & \textbf{Recall} & \textbf{Pass@1} & \textbf{F1 score} & \textbf{Precision} & \textbf{Recall} & \textbf{Pass@1} \\
    \midrule
    GPT-4.1     & \fscore{94.8}(\textcolor{Maroon!75!black}{25.7\%$\uparrow$}) & 94.2\% & 95.4\% & 95.0\% & \fscore{69.1} & 52.9\% & 99.5\% & 57.4\% \\
    GPT-4o     & \fscore{82.0}(\textcolor{Maroon!75!black}{15.8\%$\uparrow$}) & 91.2\% & 74.4\% & 84.4\% & \fscore{66.2} & 51.3\% & 93.3\% & 54.6\% \\
    DeepSeek-R1     & \fscore{83.9}(\textcolor{Maroon!75!black}{7.00\%$\uparrow$}) & 94.6\% & 75.4\% & 86.0\% & \fscore{76.9} & 84.2\% & 70.8\% & 79.6\% \\
    DeepSeek-V3.1     & \fscore{93.2}(\textcolor{Maroon!75!black}{22.0\%$\uparrow$}) & 97.0\% & 89.8\% & 93.8\% & \fscore{71.2} & 70.6\% & 71.9\% & 72.2\% \\
    Qwen3-Max     & \fscore{94.8}(\textcolor{Maroon!75!black}{23.2\%$\uparrow$}) & 98.4\% & 91.6\% & 95.2\% & \fscore{71.6} & 59.0\% & 91.0\% & 65.6\% \\
    \midrule
    GPT-4o-mini     & \fscore{37.3} & 77.2\% & 24.5\% & 60.4\% & \fscore{66.6} (\textcolor{Maroon!75!black}{29.3\%$\downarrow$}) & 51.0\% & 95.9\% & 54.1\% \\

    \bottomrule
\end{tabular}
\end{table*}

\begin{figure}
    \centering
    \includegraphics[width=\columnwidth]{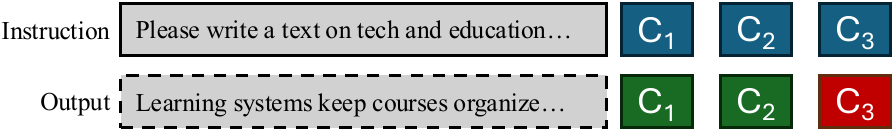}
    \vspace{-17.5pt}
    \caption{An example data entry in \bench{}}
    \vspace{-10pt}
    \label{fig:data-point}
\end{figure}

\subsection{Output Generation}

Concretely, for a template output generated by LLMs, we carefully inspect whether the output follows the instruction.
If it does not follow the instruction, we regenerate and reinspect the template output till it satisfies the instruction.
Then, we craft \texttt{unsat} outputs. We mutate the instruction in the way that the output violates the constraints in Table~\ref{tab:error}.
We balance the \texttt{sat} and \texttt{unsat} outputs in \bench{} to avoid major data skews in the dataset.
Among the 820 data points in \bench{}, 391 (47.68\%) of them have \texttt{sat} outputs and 429 (52.32\%) have \texttt{unsat} outputs.
    
Figure~\ref{fig:data-point} exemplifies data entries in \bench{}.
Every instruction in \bench{} has clear labels of its constraints; the corresponding output also has a clear result label on whether it is \texttt{sat} or \texttt{unsat}, and, if it is \texttt{unsat}, has fine-grained labels on which constraints are violated.
Therefore, \bench{} enables fine-grained evaluation on a verifier's ability to reason over multiple interacting constraints within an instruction, rather than treating instruction following as a single-label decision.

\section{Evaluation} 
\label{sec:results}

We evaluate \toolname{} and compare it with the reference verifiers on \bench{}.

\para{Reference solutions.}  
We compare \toolname{} with three LLM-as-a-judge techniques.

\begin{itemize}[itemsep=0pt]
	\item {\bf Baseline.} We use a standard implementation of LLM-as-a-judge described in~\citet{judging-llm-as-judge}; a two-shot prompt is used to show the LLM one \texttt{unsat} and one \texttt{sat} example (Appendix~\ref{sec:prompt}).
\end{itemize}

We further improve the baseline with two advanced prompt optimization techniques.

\begin{itemize}[itemsep=0pt]
	\item {\bf \gepabaseline{}} optimizes the baseline with a Chain-Of-Thought (CoT) prompt using the GEPA prompt optimization algorithm~\cite{gepa} provided by DSPy~\cite{dspy}.
	
	\item {\bf \cotbaseline{}} is a conversation-based LLM-as-a-judge technique~\cite{autogen} where multiple roles converse with each other to reach a verdict on instruction following.  
	It contains:
	\begin{itemize}[itemsep=0pt]
		\item a literalist that lists constraints in an instruction,
		\item a pragmatist who summarizes the intent behind each instruction, and
		\item the skeptic who scrutinizes the process.
	\end{itemize}
	If the roles disagree, they must debate to reach a consensus. This process enables more fine-grained CoT reasoning than basic ``think step-by-step'' CoT.
\end{itemize}

We also include a variant of \toolname{} as ablation analysis.

\begin{itemize}[itemsep=0pt]
	\item {\bf \toolnamesolver{}} is a variant of \toolname{} which only contains the SMT solver.  
	Different from \toolname{}, it directly queries LLMs to check every constraint without logical reasoning.
\end{itemize}

\para{Models.}  
To make the evaluation simple and easy to understand, we use one LLM for each evaluated verifier, instead of customizing LLMs for different components.  
We use three model families: GPT, DeepSeek, and Qwen3; specifically:  
GPT-4.1 ({\small{2025-04-14}}),  
GPT-4o ({\small{2024-11-20}}),  
DeepSeek-V3.1,  
DeepSeek-R1 ({\small{0528}}), and  
Qwen3-Max ({\small{2025-09-23}}).  
GPT-4.1 is used as the default LLM for ablation analysis in \S\ref{sec:eval:ablation} and agent-oriented study in \S\ref{sec:eval:agent}, as it is the most accessible model offered to authors.  
Although \toolname{} expects strong LLMs as it requires capabilities like formulation and code generalization, we include a small model GPT-4o-mini ({\small{2024-07-18}}) in the evaluation to understand the behavior of small models.  

\para{Metrics.}  
We measure verification effectiveness using four metrics: {\it F1 score}, {\it Precision}, {\it Recall}, and {\it Pass@1}~\citep{chen2021evaluatinglargelanguagemodels}.  
We consider F1 score as the primary metric as it balances precision and recall.

\subsection{Verification Effectiveness}
\label{sec:eval:effectiveness}

Table~\ref{tab:rq-vif} shows the effectiveness of instruction following verification of \toolname{} over the baseline using different LLMs.  
\toolname{} significantly outperforms the baseline.  
With DeepSeek-R1 as a reasoning model, the baseline achieves the highest F1 score of 76.9\%; \toolname{} achieves 90+\% F1 scores using GPT-4.1, DeepSeek-V3.1, and Qwen3-Max.  
On the same LLM, \toolname{} outperforms the baseline by up to 25.7\% increase in F1 scores.  

The baseline has higher recall than precision; we observe that LLMs tend to have more false positives than false negatives (\S\ref{sec:incompleteness}).  
Only DeepSeek models show a balance between \texttt{sat} and \texttt{unsat}.  
Some \texttt{unsat} outputs in \bench{} appear to satisfy all the constraints, but in fact violate certain logic constraints.  
For example, an output can fulfill the writing tone and topic constraints but fail to precisely hit word count.  
The baseline has a hard time catching these subtle violations.

\subsubsection{Ablation Analysis}
\label{sec:eval:ablation}

We conduct a comparative analysis of \toolname{} with \gepabaseline{}, \cotbaseline{}, and \toolnamesolver{}, which serves as an ablation study of \toolname{}.  
\gepabaseline{} and \cotbaseline{} stand for the best LLM-as-a-judge approaches without the CSP formulation.  
\toolnamesolver{} uses CSP formulation but does not do logic reasoning for individual logic constraints.

Table~\ref{tab:component-ablation} shows the results of comparative analysis, where all verifiers use GPT-4.1.  
Both \gepabaseline{} and \cotbaseline{} outperform the baseline, as the optimized prompts help LLMs think through the constraints and be more effective on checking certain constraints.  
For example, \gepabaseline{} introduces the idea of constraints and step-by-step verification into the basic CoT prompt.  
Both \gepabaseline{} and \cotbaseline{} first list all the constraints in the instruction and verify them step by step.

\begin{table}
\caption{Verification effectiveness of \toolname{} with other advanced verifiers (all using GPT-4.1). The results serve as an ablation study (\S\ref{sec:eval:ablation})}
\vspace{-5pt}
\label{tab:component-ablation}
\centering
\footnotesize
\setlength{\tabcolsep}{4.5pt}
    \begin{tabular}{lcccc}
    \toprule
    \textbf{Verifier} & \textbf{F1 Score} & \textbf{Precision} & \textbf{Recall} &  \textbf{Pass@1} \\
    \midrule
    {\gepabaseline} & \fscore{67.1} & 84.2\% & 55.8\% & 73.8\% \\
    {\cotbaseline}  & \fscore{78.9} & 73.9\% & 84.7\% & 78.4\%\\
    {\toolnamesolver} & \fscore{84.0} & 80.9\% & 87.5\% & 84.1\%\\
    {\toolname} (Full) & \fscore{94.8} & 94.2\% & 95.4\% & 95.0\% \\
    \bottomrule
\end{tabular}
\end{table}

\begin{figure}[t]
    \centering
    \includegraphics[width=0.95\columnwidth]{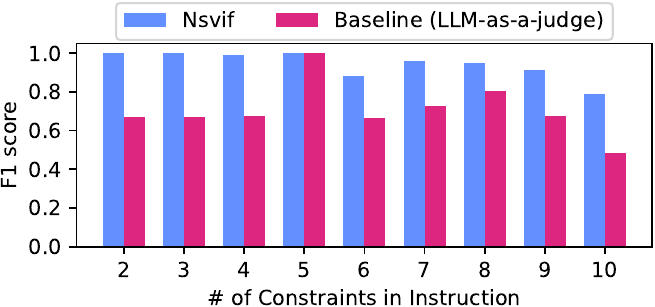}
    \vspace{-5pt}
    \caption{F1 scores of \toolname{} and the baseline by number of constraints in an instruction
    (using GPT-4.1)}
    \label{fig:correct-verification-by-num-constraints}
    \vspace{-10pt}
\end{figure}

However, they are less effective in verifying the constraints, compared with \toolnamesolver{} and \toolname{} which use SMT solvers for formal reasoning.  
As expected, leveraging symbolic reasoning can substantially improve verification effectiveness.  
Note that \toolname{} outperforms \toolnamesolver{}.  
Due to the lack of logic checkers, \toolnamesolver{} is weak on logic constraints. 
For example, we observe that it is difficult for \toolnamesolver{} to check word-count constraints; such logic constraints are challenging for LLMs as the judgments, but are straightforward for generated code.

\subsubsection{Instruction Complexity}
\label{sec:eval:complexity}

\bench{} covers instructions with different levels of complexity, in terms of the number of constraints.  
Figure~\ref{fig:correct-verification-by-num-constraints} breaks down the effectiveness of {\toolname} and the baseline (both using GPT-4.1) for different instruction complexity levels, from two to ten constraints respectively.  
We see no clear trend of effectiveness with the number of constraints.

\subsection{Impacts of LLM Capabilities}

We evaluated \toolname{} on multiple LLMs, including strong models (GPT-4.1, DeepSeek-V3.1, and Qwen3-Max) and small models (GPT-4o-mini).

Compared with GPT-4.1, GPT-4o displays more failures in parsing instructions.  
For example, GPT-4o fails to parse part of the constraint on ignoring certain lines in the output when counting words, which misleads the checking agent to generate incorrect code.  
We expect DeepSeek-R1, as a reasoning model, to perform better than DeepSeek-V3.1 when used in \toolname{}.  
However, DeepSeek-R1 displays more false-negative judgments on analyzing semantic constraints, causing lower recall.

We also evaluate \toolname{} on a small model---GPT-4o-mini.  
Unexpectedly, with GPT-4o-mini, the effectiveness of \toolname{} is worse than the baseline (Table~\ref{tab:rq-vif}).  
\toolname{} requires the LLM to be capable of:  
(1) identifying the constraints,  
(2) classifying every constraint as logic or semantic, and  
(3) generating checker code for logic constraints.  
GPT-4o-mini is effective in (1)---among 820 instructions, GPT-4o-mini correctly identifies all constraints in 734 instructions.  
However, it misclassifies constraint(s) in 508 instructions, e.g., misclassifying writing topics as logic constraints, forcing the checking agent to generate code to check them. For semantic constraints, it is hard for GPT-4o-mini to generate effective code, e.g., it generates a simple keyword check (e.g., if \texttt{education} is in the text) for topic constraints.  
We did not expect misclassification to be a bottleneck and lead to negative results.

\begin{table}
\caption{The percentage (number) of false positives and false negatives of 
    each approach with GPT-4.1}
\label{tab:fn-fp}
\vspace{-5pt}
\centering
\setlength{\tabcolsep}{8pt}
\footnotesize
\begin{tabular}{lcc}
\toprule
\textbf{Verifier} & \textbf{FP \% (\#)} & \textbf{FN \% (\#)} \\
\midrule
Baseline & 42.32\% (347) & 0.24\% (2) \\
\gepabaseline{} & 5.00\% (41) & 21.10\% (173) \\
\cotbaseline{} & 14.27\% (117) & 7.32\% (60) \\
\toolnamesolver{} & 9.88\% (81) & 5.98\% (49) \\
\toolname{} (Full) & 2.80\% (23) & 2.20\% (18) \\
\bottomrule
\end{tabular}
\end{table}

\subsection{Soundness and Completeness}
\label{sec:incompleteness}

We analyzed the soundness and completeness of \toolname{} as a verifier.  
Soundness is measured by false positives and completeness is measured by false negatives.  
We define a false-positive verification result as the verifier producing a \texttt{sat} result when the output does not satisfy the instruction, and a false-negative verification result as the verifier producing an \texttt{unsat} result when the output actually satisfies the instruction.
Table~\ref{tab:fn-fp} shows the two metrics of \toolname{} and the reference verifiers (all using GPT-4.1).  
All the references have high rates (around 10\%) of either false positives or false negatives.  
We observe that GPT-4.1 has trouble checking logic constraints such as word counts (either over- or under-counting).  
In fact, GPT-4.1 tends to estimate the word count, instead of counting it precisely.  
Moreover, GPT-4.1 struggles to correctly solve first-order logic, resulting in incorrect verdict.

\toolnamesolver{} outperforms \gepabaseline{} and \cotbaseline{} because it takes the step of writing the Z3 program.  
Certainly, we observe that GPT-4.1 occasionally generates incorrect formula, causing both false positives and false negatives.

\toolname{} has much fewer false positives and false negatives.  
However, it still suffers from hallucination behavior of GPT-4.1.  
For the 23 false positives, GPT-4.1 hallucinates and modifies the given output when invoking the checker instead of copying the original text.  
The hallucinated output satisfies the instruction, causing \toolname{} to output \texttt{sat} when the ground-truth is \texttt{unsat}.  
For the 18 false negatives:  
(1) 12 were caused by GPT-4.1 generating wrong results against semantic constraints;  
(2) 4 were caused by checker generation failures and fallback to LLM-as-a-judge which produced wrong results;  
(3) one was caused by the formulation agent incorrectly extracting constraints (introducing a constraint that was not in the original instruction); and  
(4) one was caused by the hallucination behavior similar as the false positives.  
We believe some of them could be further reduced by patching our current implementation of \toolname{}.

\subsection{Multi-turn Feedback}
\label{sec:eval:agent}

\begin{figure}[t]
\centering
\includegraphics[width=\columnwidth]{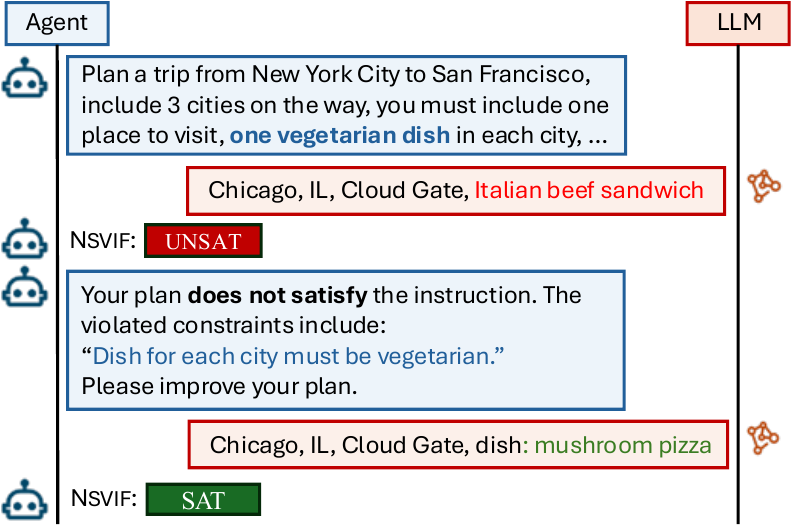}
\vspace{-10pt}
\caption{Workflow of \toolname{} improving LLM instruction following via multi-turn feedback.}
\vspace{-10pt}
\label{fig:travel-agent}
\end{figure}

To show that \toolname{} can help LLMs follow instructions in agentic settings, we write a simple travel agent that prompts GPT-4.1 to generate a travel plan with three constraints (see Figure~\ref{fig:travel-agent}).  
GPT-4.1's initial output violates the dietary constraint.  
The agent uses \toolname{} to pinpoint the violation and ask GPT-4.1 to regenerate the plan with the violation information, until it passes \toolname{}.  
GPT-4.1 takes one iteration to fix the dish item in the output and produce a \texttt{sat} output.

We further selected nine instructions from \bench{} which the LLM (GPT-4.1) failed to follow.  
The nine instructions have different numbers of constraints from 2 to 10, sampled from the complexity group in Figure~\ref{fig:correct-verification-by-num-constraints}.  
We evaluate (1) how \toolname{}'s output can help GPT-4.1 iteratively improve its output and (2) how many iterations GPT-4.1 takes to reach an instruction-following output with \toolname{} feedback.  
For all the instructions, GPT-4.1 eventually produces a \texttt{sat} output in less than 5 iterations when guided by \toolname{}; 5 (55.6\%) are corrected in less than 3 iterations.  
Difficult constraints include total word count and writing tone, which take GPT-4.1 several iterations to satisfy.  
For the word-count constraint, the word count of the initial output greatly violates the constraint; GPT-4.1 iteratively improves the output by editing sentences while still keeping the text on topic.  
This shows that even though the LLM is not capable of symbolic reasoning, given the right feedback, it can still follow the feedback and improve upon itself.  
For the writing tone constraint, GPT-4.1 tunes words and gradually makes text pass verification.

We then experimented with a boolean feedback similar to the baseline approach.  
When the output violates the instruction, we only give GPT-4.1 the single \texttt{unsat} result and ask it to regenerate.  
We give GPT-4.1 up to 15 iterations to generate a satisfying output.  
We observe that for 3 instructions, GPT-4.1 cannot generate a satisfying output within the budget, falling into a loop of retrying.  
One instruction took GPT-4.1 11 iterations to finally produce a satisfying output.  
For the remaining 5 instructions, GPT-4.1 took less than three iterations to correct the output.  
The experiment shows the usefulness of \toolname{}' feedback in improving instruction-following ability of LLMs without additional post-training.

\section{Related Work}
\label{sec:related}

Like \toolname{}, prior work also models instructions to LLMs as constraints~\cite{complexbench,IFEval,jiang-etal-2024-followbench}.  
However, as the focus was on benchmarking the instruction-following ability of LLMs, prior work uses predefined sets of constraints and designs corresponding validators {\it a priori}.  
\toolname{} addresses a different problem. It serves as a universal verifier of instruction following without assuming the shapes of input instruction or output.  
It shares the idea of decomposing an instruction into constraints~\cite{complexbench,qin2024infobenchevaluatinginstructionfollowing}; it does so by formulating instruction following as a constraint-satisfaction problem, with both logic and semantic constraints which can be unified in first-order logic.  
\toolname{} shows that bringing symbolic reasoning for logic constraints substantially improves its verification effectiveness.

We discussed the limitations of existing benchmarks for evaluating instruction following verification, namely LLMBar~\citep{zeng2024evaluatinglargelanguagemodels} and ReIFE~\cite{reife} in \S\ref{sec:bench}.  
In short, these benchmarks primarily rely on preference-based judgments with under-specified acceptance criteria, making it difficult to have fine-grained analysis or comprehend the verifier behavior.  
\bench{} is inspired by benchmarks for evaluating instruction following of LLMs such as~\citet{complexbench}.  
Note that such LLM instruction following benchmarks do not apply directly, because they only contain the instructions but do not have labeled outputs.

\section{Concluding Remarks}
\label{sec:conclusion}

In search of a more principled technique than LLMs as verifiers, we explored a neuro-symbolic approach that integrates logic reasoning to verify instruction-following behavior of LLMs and shows its promises.  
\toolname{} serves as a first step and a framework to enable more specialized and personalized verifiers and ensure instruction following of modern LLMs.

\section{Limitations}
\label{sec:limitation}

On problem formulation, constraint-satisfaction problem (CSP) is not the only way to formulate instruction-following verification.  
Specifically, CSP does not differentiate soft constraints and hard constraints.  
It is possible that a given instruction is unsatisfiable, and unsatisfiable instructions could lead to infinite retries or task failures.  
We plan to explore a new formulation of {\it constraint-optimization problem} or {\it COP}, which is a generalization of CSP.  
The challenge is to autonomously decide the objective function to be optimized.

As an LLM-powered tool, \toolname{} does not escape from the fundamental paradox---using LLMs to verify instruction following of LLMs.  
It is challenging to completely eliminate hallucination and nondeterminism behavior inherent in existing LLMs during \toolname{}'s verification.  
On the other hand, the neuro-symbolic approach of \toolname{} minimizes the impact of LLM's nondeterminism by:  
1. enforcing LLMs to focus on specific, well-defined tasks such as classification and code generation, and  
2. leveraging symbolic reasoning at critical steps such as verifying individual constraints with programs and finally solving the constraints with a solver.

As a benchmark, \bench{} is currently limited to English writing tasks.  
Writing tasks are common in evaluating instruction-following ability of LLMs~\citep{IFEval, collie, qin2024infobenchevaluatinginstructionfollowing, li-etal-2024-evaluating-instruction}.  
It currently does not consider multilingual instructions~\citep{dussolle-etal-2025-ifeval}.  
Our future work considers extending \bench{} to tasks beyond writing tasks, such as information retrieval, general question answering, and planning.

Our evaluation was limited to the LLMs that we have access to and have the budget to run.  
Other LLMs could manifest different characteristics and may lead to different results.  
On the other hand, we believe that the fundamental problem is rooted across all commodity LLMs, and our approach is generally applicable to other LLMs that are not evaluated in this work.

\bibliography{main}

@inproceedings{zhang2025agent,
 author = {Zhang, Shaokun and Yin, Ming and Zhang, Jieyu and Liu, Jiale and Han, Zhiguang and Zhang, Jingyang and Li, Beibin and Wang, Chi and Wang, Huazheng and Chen, Yiran and Wu, Qingyun},
 booktitle = {Proceedings of The 42nd International Conference on Machine Learning (ICML'25)},
 title = "{Which Agent Causes Task Failures and When? On Automated Failure Attribution of LLM Multi-Agent Systems}",
 year = {2025},
 month = {MAY},
 doi = {10.48550/arXiv.2505.00212},
 url = {https://openreview.net/forum?id=GazlTYxZss},
}

@inproceedings{cemri2025multi,
 author = {Mert Cemri and Melissa Z. Pan and Shuyi Yang and Lakshya A. Agrawal and Bhavya Chopra and Rishabh Tiwari and Kurt Keutzer and Aditya Parameswaran and Dan Klein and Kannan Ramchandran and Matei Zaharia and Joseph E. Gonzalez and Ion Stoica},
 booktitle = {Proceedings of The 39th Annual Conference on Neural Information Processing Systems (NeurIPS'25)},
 doi = {10.48550/arXiv.2503.13657},
 title = "{Why Do Multi-Agent LLM Systems Fail?}",
 year = {2025},
 month = {MAR}
}

@article{qin2024infobenchevaluatinginstructionfollowing,
 author = {Yiwei Qin and Kaiqiang Song and Yebowen Hu and Wenlin Yao and Sangwoo Cho and Xiaoyang Wang and Xuansheng Wu and Fei Liu and Pengfei Liu and Dong Yu},
 journal = {arXiv:2401.03601},
 title = "{InFoBench: Evaluating Instruction Following Ability in Large Language Models}",
 year = {2024},
 month={JAN},
 url = {https://arxiv.org/abs/2401.03601},
}

@inproceedings{jiang-etal-2024-followbench,
    title = "{FollowBench: A Multi-level Fine-grained Constraints Following Benchmark for Large Language Models}",
    author = "Jiang, Yuxin  and
      Wang, Yufei  and
      Zeng, Xingshan  and
      Zhong, Wanjun  and
      Li, Liangyou  and
      Mi, Fei  and
      Shang, Lifeng  and
      Jiang, Xin  and
      Liu, Qun  and
      Wang, Wei",
    booktitle = "Proceedings of the 62nd Annual Meeting of the Association for Computational Linguistics (ACL'24)",
    month = aug,
    year = 2024,
    url = "https://aclanthology.org/2024.acl-long.257/",
    doi = "10.18653/v1/2024.acl-long.257",
}

@inproceedings{dussolle-etal-2025-ifeval,
    title = "{{M-IFEval: Multilingual Instruction-Following Evaluation}}",
    author = "Dussolle, Antoine  and
      Carde{\~n}a, A.  and
      Sato, Shota  and
      Devine, Peter",
    editor = "Chiruzzo, Luis  and
      Ritter, Alan  and
      Wang, Lu",
    booktitle = "Proceedings of the 2025 Annual Conference of the Nations of the Americas Chapter of the 
        Association for Computational Linguistics (NAACL'25)",
    month = apr,
    year = "2025",
    url = "https://aclanthology.org/2025.findings-naacl.344/",
    doi = "10.18653/v1/2025.findings-naacl.344",
    ISBN = "979-8-89176-195-7",
}

@inproceedings{yuan-etal-2025-following,
    title = "{Following Length Constraints in Instructions}",
    author = "Yuan, Weizhe  and
      Kulikov, Ilia  and
      Yu, Ping  and
      Cho, Kyunghyun  and
      Sukhbaatar, Sainbayar  and
      Weston, Jason E  and
      Xu, Jing",
    booktitle = "Proceedings of the 2025 Conference on Empirical Methods in Natural Language Processing (EMNLP'25)",
    month = nov,
    year = "2025",
    url = "https://aclanthology.org/2025.emnlp-main.1233/",
    doi = "10.18653/v1/2025.emnlp-main.1233",
}

@inproceedings{pham-etal-2024-suri,
    title = "{Suri: Multi-constraint Instruction Following in Long-form Text Generation}",
    author = "Pham, Chau Minh  and
      Sun, Simeng  and
      Iyyer, Mohit",
    editor = "Al-Onaizan, Yaser  and
      Bansal, Mohit  and
      Chen, Yun-Nung",
    booktitle = "Findings of the Association for Computational Linguistics: EMNLP 2024 (EMNLP'24)",
    month = nov,
    year = "2024",
    address = "Miami, Florida, USA",
    publisher = "Association for Computational Linguistics",
    url = "https://aclanthology.org/2024.findings-emnlp.94/",
    doi = "10.18653/v1/2024.findings-emnlp.94",
    pages = "1722--1753",
}

@article{IFEval,
 author = {Jeffrey Zhou and Tianjian Lu and Swaroop Mishra and Siddhartha Brahma and Sujoy Basu and Yi Luan and Denny Zhou and Le Hou},
 journal = {arXiv:2311.07911},
 title = "{Instruction-Following Evaluation for Large Language Models}",
 year = {2023},
 month={NOV},
 doi = {10.48550/arXiv.2311.07911},
 url = {https://arxiv.org/abs/2311.07911},
}

@inproceedings{he2024largelanguagemodelsunderstand,
 author = {Qianyu He and
Jie Zeng and
Wenhao Huang and
Lina Chen and
Jin Xiao and
Qianxi He and
Xunzhe Zhou and
Jiaqing Liang and
Yanghua Xiao},
 booktitle = {Proceedings of the 38th AAAI Conference on Artificial Intelligence (AAAI'24)},
 title = "{Can Large Language Models Understand Real-World Complex Instructions?}",
 url = {https://doi.org/10.1609/aaai.v38i16.29777},
 year = {2024}
}

@article{qi2025agentifbenchmarkinginstructionfollowing,
 author = {Yunjia Qi and Hao Peng and Xiaozhi Wang and Amy Xin and Youfeng Liu and Bin Xu and Lei Hou and Juanzi Li},
 journal = {arXiv:2505.16944},
 title = "{AGENTIF: Benchmarking Instruction Following of Large Language Models in Agentic Scenarios}",
 doi = {10.48550/arXiv.2505.16944},
 year = {2025}
}

@inproceedings{oasst1,
author = {K\"{o}pf, Andreas and Kilcher, Yannic and von R\"{u}tte, Dimitri and Anagnostidis, Sotiris and Tam, Zhi-Rui and Stevens, Keith and Barhoum, Abdullah and Duc, Nguyen Minh and Stanley, Oliver and Nagyfi, Rich\'{a}rd and ES, Shahul and Suri, Sameer and Glushkov, David and Dantuluri, Arnav and Maguire, Andrew and Schuhmann, Christoph and Nguyen, Huu and Mattick, Alexander},
title = "{OpenAssistant conversations - democratizing large language model alignment}",
year = {2023},
doi = {10.5555/3666122.3668186},
booktitle = {Proceedings of the 37th Annual Conference on Neural Information Processing Systems (NeurIPS'23)},
month = {DEC},
}

@inproceedings{lmsys1m,
      title="{LMSYS-Chat-1M: A Large-Scale Real-World LLM Conversation Dataset}",
      booktitle={Proceedings of The 12th International Conference on Learning Representations (ICLR'24)},
      author={Lianmin Zheng and Wei-Lin Chiang and Ying Sheng and Tianle Li and Siyuan Zhuang and Zhanghao Wu and Yonghao Zhuang and Zhuohan Li and Zi Lin and Eric P. Xing and Joseph E. Gonzalez and Ion Stoica and Hao Zhang},
      year={2024},
      month={JAN},
      url={https://openreview.net/forum?id=BOfDKxfwt0},
}

@inproceedings{wildchat,
  author       = {Wenting Zhao and
                  Xiang Ren and
                  Jack Hessel and
                  Claire Cardie and
                  Yejin Choi and
                  Yuntian Deng},
  title        = "{WildChat: 1M ChatGPT Interaction Logs in the Wild}",
  booktitle    = {Proceedings of The 12th International Conference on Learning Representations (ICLR'24)},
  year         = {2024},
  url          = {https://openreview.net/forum?id=Bl8u7ZRlbM},
  month = {MON},
}

@article{sirdeshmukh2025multichallengerealisticmultiturnconversation,
 author = {Ved Sirdeshmukh and Kaustubh Deshpande and Johannes Mols and Lifeng Jin and Ed-Yeremai Cardona and Dean Lee and Jeremy Kritz and Willow Primack and Summer Yue and Chen Xing},
 journal = {arXiv:2501.17399},
 title = "{MultiChallenge: A Realistic Multi-Turn Conversation Evaluation Benchmark Challenging to Frontier LLMs}",
 year = {2025},
 month = {JAN},
 doi = {10.18653/v1/2025.findings-acl.958}
}

@inproceedings{zeng2024evaluatinglargelanguagemodels,
 author = {Zhiyuan Zeng and
Jiatong Yu and
Tianyu Gao and
Yu Meng and
Tanya Goyal and
Danqi Chen},
 booktitle = {Proceedings of The 12th International Conference on Learning Representations (ICLR'24)},
 title = "{Evaluating Large Language Models at Evaluating Instruction Following}",
 year = {2024},
 month = {DEC},
 doi = {10.48550/arXiv.2310.07641},
 url = {https://openreview.net/forum?id=tr0KidwPLc}
}

@article{wu2025lifbenchevaluatinginstructionfollowing,
 author = {Xiaodong Wu and Minhao Wang and Yichen Liu and Xiaoming Shi and He Yan and Xiangju Lu and Junmin Zhu and Wei Zhang},
 journal = {arXiv:2411.07037},
 title = "{LIFBench: Evaluating the Instruction Following Performance and Stability of Large Language Models in Long-Context Scenarios}",
 doi = {10.48550/arXiv.2411.07037},
 year = {2024}
}

@inproceedings{judging-llm-as-judge,
 author = {Lianmin Zheng and
Wei{-}Lin Chiang and
Ying Sheng and
Siyuan Zhuang and
Zhanghao Wu and
Yonghao Zhuang and
Zi Lin and
Zhuohan Li and
Dacheng Li and
Eric P. Xing and
Hao Zhang and
Joseph E. Gonzalez and
Ion Stoica},
 booktitle = {Proceedings of the 37th Annual Conference on Neural Information Processing Systems (NeurIPS'23)},
 title = "{Judging LLM-as-a-Judge with MT-Bench and Chatbot Arena}",
 year = {2023},
 doi = {10.5555/3666122.3668142},
 month={DEC}
}

@article{deepseek-v3.1,
 author = {DeepSeek-AI},
 journal = {arXiv:2412.19437},
 title = "{DeepSeek-V3 Technical Report}",
 url = {https://arxiv.org/abs/2412.19437},
 year = {2024}
}

@article{laban2025llmslostmultiturnconversation,
 author = {Philippe Laban and Hiroaki Hayashi and Yingbo Zhou and Jennifer Neville},
 journal = {arXiv:2505.06120},
 title = "{LLMs Get Lost In Multi-Turn Conversation}",
 doi = {10.48550/arXiv.2505.06120},
 year = {2025},
 month = {MAY},
}

@article{jaroslawicz2025instructionsllmsfollowonce,
 author = {Daniel Jaroslawicz and Brendan Whiting and Parth Shah and Karime Maamari},
 journal = {arXiv:2507.11538},
 title = "{How Many Instructions Can LLMs Follow at Once?}",
 doi = {10.48550/arXiv.2507.11538},
 year = {2025},
 month = {JUL}
}

@misc{openai2025gpt5,
 author = {OpenAI},
 title = "{Introducing GPT-5}",
 howpublished = {\url{https://openai.com/index/introducing-gpt-5/}},
 year = {2025}
}

@article{chen2021evaluatinglargelanguagemodels,
 author = {Mark Chen and Jerry Tworek and Heewoo Jun and Qiming Yuan and Henrique Ponde de Oliveira Pinto and Jared Kaplan and Harri Edwards and Yuri Burda and Nicholas Joseph and Greg Brockman and Alex Ray and Raul Puri and Gretchen Krueger and Michael Petrov and Heidy Khlaaf and Girish Sastry and Pamela Mishkin and Brooke Chan and Scott Gray and Nick Ryder and Mikhail Pavlov and Alethea Power and Lukasz Kaiser and Mohammad Bavarian and Clemens Winter and Philippe Tillet and Felipe Petroski Such and Dave Cummings and Matthias Plappert and Fotios Chantzis and Elizabeth Barnes and Ariel Herbert-Voss and William Hebgen Guss and Alex Nichol and Alex Paino and Nikolas Tezak and Jie Tang and Igor Babuschkin and Suchir Balaji and Shantanu Jain and William Saunders and Christopher Hesse and Andrew N. Carr and Jan Leike and Josh Achiam and Vedant Misra and Evan Morikawa and Alec Radford and Matthew Knight and Miles Brundage and Mira Murati and Katie Mayer and Peter Welinder and Bob McGrew and Dario Amodei and Sam McCandlish and Ilya Sutskever and Wojciech Zaremba},
 journal = {arXiv:2107.03374},
 title = "{Evaluating Large Language Models Trained on Code}",
 doi = {10.48550/arXiv.2107.03374},
 year = {2021}
}

@inproceedings{lmql,
author = {Beurer-Kellner, Luca and Fischer, Marc and Vechev, Martin},
title = "{Prompting Is Programming: A Query Language for Large Language Models}",
doi = {10.1145/3591300},
booktitle = {Proceedings of the 44th ACM SIGPLAN Conference on Programming Language Design and Implementation (PLDI'23)},
year = {2023},
month = {JUN},
}

@inproceedings{dspy,
  title="{DSPy: Compiling Declarative Language Model Calls into Self-Improving Pipelines}",
  doi = {10.48550/arXiv.2310.03714},
  author={Khattab, Omar and Singhvi, Arnav and Maheshwari, Paridhi and Zhang, Zhiyuan and Santhanam, Keshav and Vardhamanan, Sri and Haq, Saiful and Sharma, Ashutosh and Joshi, Thomas T. and Moazam, Hanna and Miller, Heather and Zaharia, Matei and Potts, Christopher},
  booktitle={Proceedings of the 12th International Conference on Learning Representations (ICLR'24)},
  year={2024},
  month={JAN}
}

@article{gepa,
      title="{GEPA: Reflective Prompt Evolution Can Outperform Reinforcement Learning}", 
      author={Lakshya A Agrawal and Shangyin Tan and Dilara Soylu and Noah Ziems and Rishi Khare and Krista Opsahl-Ong and Arnav Singhvi and Herumb Shandilya and Michael J Ryan and Meng Jiang and Christopher Potts and Koushik Sen and Alexandros G. Dimakis and Ion Stoica and Dan Klein and Matei Zaharia and Omar Khattab},
      journal={arXiv:2507.19457},
      year={2025},
      month={JUL},
      doi={10.48550/arXiv.2507.19457},
}

@inproceedings{autogen,
      title="{AutoGen: Enabling Next-Gen LLM Applications via Multi-Agent Conversation}", 
      author={Qingyun Wu and Gagan Bansal and Jieyu Zhang and Yiran Wu and Beibin Li and Erkang Zhu and Li Jiang and Xiaoyun Zhang and Shaokun Zhang and Jiale Liu and Ahmed Hassan Awadallah and Ryen W White and Doug Burger and Chi Wang},
      booktitle={Proceedings of the 1st Conference on Language Modeling (COLM'24)},
      year = {2024},
      month = {AUG},
      url = {https://openreview.net/forum?id=BAakY1hNKS}
}

@article{pdl,
      title="{PDL: A Declarative Prompt Programming Language}", 
      author={Mandana Vaziri and Louis Mandel and Claudio Spiess and Martin Hirzel},
      journal={arXiv:2410.19135},
      year={2024},
      month={OCT},
      doi = {10.48550/arXiv.2410.19135},
}

@book{Albarghouthi2021,
author = {Albarghouthi, Aws},
title = "{Introduction to Neural Network Verification}",
doi = {10.1561/2500000051},
year = {2021},
issue_date = {Dec 2021},
publisher = {Now Publishers Inc.},
volume = {7},
number = {1–2},
month = dec,
pages = {1–157},
numpages = {164}
}

@inproceedings{z3,
  author = {De Moura, Leonardo and Bj\o{}rner, Nikolaj},
  title = "{Z3: An Efficient SMT Solver}",
  year = {2008},
  month = {MAR},
  doi = {10.1007/978-3-540-78800-3_24},
  booktitle = {Proceedings of the 14th International Conference on Tools and Algorithms for the Construction and Analysis of Systems (TACAS'08)},
}

@article{ifsurvey,
    author = {Lou, Renze and Zhang, Kai and Yin, Wenpeng},
    title = "{Large Language Model Instruction Following: A Survey of Progresses and Challenges}",
    journal = {Computational Linguistics},
    volume = {50},
    number = {3},
    pages = {1053-1095},
    year = {2024},
    month = {09},
    issn = {0891-2017},
    doi = {10.1162/coli_a_00523},
}

@inproceedings{reife,
      title="{ReIFE: Re-evaluating Instruction-Following Evaluation}", 
      author={Yixin Liu and Kejian Shi and Alexander Fabbri and Yilun Zhao and PeiFeng Wang and Chien-Sheng Wu and Shafiq Joty and Arman Cohan},
      booktitle={Proceedings of the 2025 Conference of the Nations of the Americas Chapter of the Association for Computational Linguistics (ACL'25)},
      year={2025},
      month={APR},
      doi="10.18653/v1/2025.naacl-long.610",
}

@inproceedings{EIFBENCH,
      title="{EIFBENCH: Extremely Complex Instruction Following Benchmark for Large Language Models}", 
      author={Tao Zou and Xinghua Zhang and Haiyang Yu and Minzheng Wang and Fei Huang and Yongbin Li},
      booktitle={Proceedings of the 2025 Conference on Empirical Methods in Natural Language Processing (EMNLP'25)},
      year={2025},
      doi={10.18653/v1/2025.emnlp-main.1059},
      month={APR},
}

@inproceedings{geval,
      title="{G-Eval: NLG Evaluation using Gpt-4 with Better Human Alignment}", 
      author={Yang Liu and Dan Iter and Yichong Xu and Shuohang Wang and Ruochen Xu and Chenguang Zhu},
      booktitle={Proceedings of the 2023 Conference on Empirical Methods in Natural Language Processing (EMNLP'23)},
      year={2023},
      month={DEC},
      doi = {10.18653/v1/2023.emnlp-main.153}
}

@inproceedings{mellea,
    title = "{From Agent Soup to Proper Software Design: Putting the Developer Back in Control of Generative AI with Mellea}",
    author = {David Cox},
    year = {2025},
    month = DEC,
    url = {https://neurips.cc/virtual/2025/talk/127761},
    booktitle = {Proceedings of the 39th Annual Conference on Neural Information Processing Systems (NeurIPS'25)},
}

@inproceedings{ifbench,
    title = "{Generalizing Verifiable Instruction Following}",
    author = {Valentina Pyatkin and Saumya Malik and Victoria Graf and 
        Hamish Ivison and Shengyi Huang and Pradeep Dasigi and Nathan Lambert and Hannaneh Hajishirzi},
    year = {2025},
    month = DEC,
    doi = {10.48550/arXiv.2507.02833},
    booktitle = {Proceedings of the 39th Annual Conference on Neural Information Processing Systems (NeurIPS'25)},
}

@inproceedings{complexbench,
    title = "{Benchmarking Complex Instruction-Following with Multiple Constraints Composition}",
    author = {Bosi Wen and Pei Ke and Xiaotao Gu and Lindong Wu and Hao Huang and Jinfeng Zhou and Wenchuang Li and Binxin Hu
        and Wendy Gao and Jiaxin Xu and Yiming Liu and Jie Tang and Hongning Wang and Minlie Huang},
    year = {2024},
    month = DEC,
    booktitle = {Proceedings of the 38th Annual Conference on Neural Information Processing Systems (NeurIPS'24)},
    doi = {10.5555/3737916.3742287},
}

@article{gao-etal-2025-llm,
    title = "{LLM-based NLG Evaluation: Current Status and Challenges}",
    author = "Gao, Mingqi  and
      Hu, Xinyu  and
      Yin, Xunjian  and
      Ruan, Jie  and
      Pu, Xiao  and
      Wan, Xiaojun",
    journal = "Computational Linguistics",
    volume = "51",
    month = jun,
    year = "2025",
    address = "Cambridge, MA",
    publisher = "MIT Press",
    url = "https://aclanthology.org/2025.cl-2.9/",
    doi = "10.1162/coli_a_00561",
    pages = "661--687",
}

@inproceedings{linc,
    author={Olausson, Theo and Gu, Alex and Lipkin, Ben and Zhang, Cedegao and Solar-Lezama, Armando and Tenenbaum, Joshua and Levy, Roger},
    title="{LINC: A Neurosymbolic Approach for Logical Reasoning by Combining Language Models with First-Order Logic Provers}",
    booktitle = "Proceedings of the 2023 Conference on Empirical Methods in Natural Language Processing (EMNLP'23)",
    month = dec,
    year = "2023",
    doi = "10.18653/v1/2023.emnlp-main.313",
}

@article{if256,
    author={Richard J. Young and Brandon Gillins and Alice M. Matthews},
    title="{When Models Can't Follow: Testing Instruction Adherence Across 256 LLMs}",
    journal = {arXiv:2510.18892},
    year = {2025},
    month = {OCT},
    doi = {10.48550/arXiv.2510.18892},
}

@inproceedings{logic-fm,
    title="{Logic-LM: Empowering Large Language Models with Symbolic Solvers for Faithful Logical Reasoning}", 
    author={Liangming Pan and Alon Albalak and Xinyi Wang and William Yang Wang},
    booktitle = "Findings of the Association for Computational Linguistics: EMNLP 2023 (EMNLP'23)",
    month = dec,
    year = "2023",
    doi = "10.18653/v1/2023.findings-emnlp.248",
}

@article{ibm-formal-agent-spec,
    title="{Formally Specifying the High-Level Behavior of LLM-Based Agents}", 
    author={Maxwell Crouse and Ibrahim Abdelaziz and Ramon Astudillo and Kinjal Basu and Soham Dan and Sadhana Kumaravel and Achille Fokoue and Pavan Kapanipathi and Salim Roukos and Luis Lastras},
    year={2023},
    month={OCT},
    doi = {10.48550/arXiv.2310.08535},
    journal={arXiv:2310.08535},
}

@article{formal-llm,
  author       = {Zelong Li and
                  Wenyue Hua and
                  Hao Wang and
                  He Zhu and
                  Yongfeng Zhang},
  title        = "{Formal-LLM: Integrating Formal Language and Natural Language for Controllable
                  LLM-based Agents}",
  journal      = {arXiv:2402.00798},
  year         = {2024},
  month        = {FEB},
  doi          = {10.48550/ARXIV.2402.00798},
}

@inproceedings{jimenez2024swebench,
title="{SWE-bench: Can Language Models Resolve Real-world Github Issues?}",
author={Carlos E Jimenez and John Yang and Alexander Wettig and Shunyu Yao and Kexin Pei and Ofir Press and Karthik R Narasimhan},
doi={10.48550/arXiv.2310.06770},
booktitle={Proceedings of The Twelfth International Conference on Learning Representations (ICLR'24)},
year={2024},
month={JAN}
}

@inproceedings{li-etal-2024-evaluating-instruction,
  title = "{Evaluating the Instruction-Following Robustness of Large Language Models to Prompt Injection}",
  author = {Li, Zekun  and
    Peng, Baolin  and
    He, Pengcheng  and
    Yan, Xifeng},
  booktitle = {Proceedings of the 2024 Conference on Empirical Methods in Natural Language Processing (EMNLP'24)},
  doi = {10.18653/v1/2024.emnlp-main.33},
  month = {NOV},
  year = {2024},
}

@inproceedings{collie,
      title="{COLLIE: Systematic Construction of Constrained Text Generation Tasks}", 
      author={Shunyu Yao and Howard Chen and Austin W. Hanjie and Runzhe Yang and Karthik Narasimhan},
      year={2024},
      month={JAN},
      booktitle={Proceedings of the 12th International Conference on Learning Representations (ICLR'24)},
      url={https://openreview.net/forum?id=kxgSlyirUZ}, 
}

@inproceedings{carp,
    title = "{Text Classification via Large Language Models}",
    author = {Sun, Xiaofei  and
      Li, Xiaoya  and
      Li, Jiwei  and
      Wu, Fei  and
      Guo, Shangwei  and
      Zhang, Tianwei  and
      Wang, Guoyin},
    booktitle = {Findings of the Association for Computational Linguistics: EMNLP 2023 (EMNLP'23)},
    month = {DEC},
    year = {2023},
    doi = {10.18653/v1/2023.findings-emnlp.603},
}

@inproceedings{position-fm,
title="{Trustworthy AI Agents Require the Integration of Large Language Models and Formal Methods}",
author={Yedi Zhang and Yufan Cai and Xinyue Zuo and Xiaokun Luan and Kailong Wang and Zhe Hou and Yifan Zhang and Zhiyuan Wei and Meng Sun and Jun Sun and Jing Sun and Jin Song Dong},
booktitle={Proceedings of 42nd International Conference on Machine Learning (ICML'25)},
year={2025},
url={https://openreview.net/forum?id=wkisIZbntD}
}

@misc{swebench-verified,
  title="{Introducing {SWE}-bench Verified}",
  author={Chowdhury, Neil and Aung, James and Shern, Chan Jun and Jaffe, Oliver and Sherburn, Dane and Starace, Giulio and Mays, Evan and Dias, Rachel and Aljubeh, Marwan and Glaese, Mia and Jimenez, Carlos E. and Yang, John and Ho, Leyton and Patwardhan, Tejal and Liu, Kevin and Madry, Aleksander},
  year={2024},
  howpublished={\url{https://openai.com/index/introducing-swe-bench-verified/}},
}

@inproceedings{specgen,
author = {Ma, Lezhi and Liu, Shangqing and Li, Yi and Xie, Xiaofei and Bu, Lei},
title = "{SpecGen: Automated Generation of Formal Program Specifications via Large Language Models}",
year = {2025},
month = {SEP},
doi = {10.1109/ICSE55347.2025.00129},
booktitle = {Proceedings of the IEEE/ACM 47th International Conference on Software Engineering (ICSE'25)},
}

@article{yang2025swesmith,
  title="{SWE-smith: Scaling Data for Software Engineering Agents}", 
  author={John Yang and Kilian Lieret and Carlos E. Jimenez and Alexander Wettig and Kabir Khandpur and Yanzhe Zhang and Binyuan Hui and Ofir Press and Ludwig Schmidt and Diyi Yang},
  year={2025},
  doi = {10.48550/arXiv.2504.21798},
  journal={arXiv:2504.21798},
}

\newpage
\appendix
\section{Details of \bench{}}

Table~\ref{tab:vifbench-const-table} provides detailed descriptions of the constraints used in \bench{} and their representation in the instructions.  
The benchmark contains 820 instructions on freeform English writing tasks.  
Each instruction contains 2 to 10 constraints from Table~\ref{tab:vifbench-const-table}.

\subsection{Constraints in \bench{}}

\begin{figure}[h!]
    \centering
    \includegraphics[width=\columnwidth]{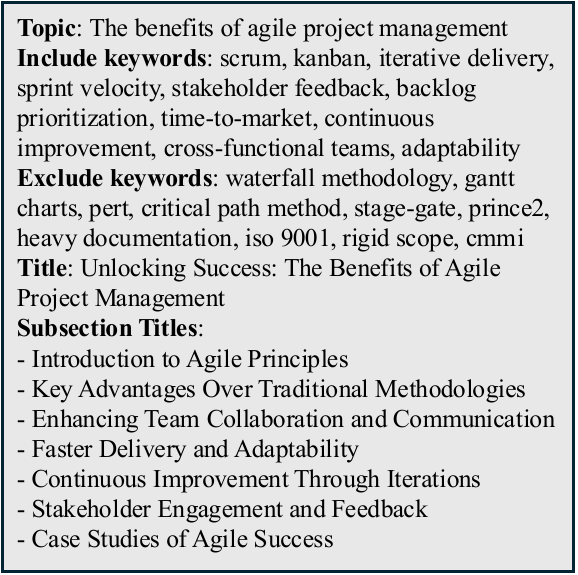}
    \caption{Example of grouped constraints for the writing topic ``The benefits of agile project management''.}
    \label{fig:const-example}
\end{figure}

Constraint types in \bench{} are sourced from instruction-turning datasets and instruction-following evaluation benchmarks~\cite{oasst1, lmsys1m, wildchat, complexbench, IFEval, qin2024infobenchevaluatinginstructionfollowing}.  
For each type of constraint, we use GPT-4.1 to generate constraint values.  
For example, for the writing topic constraint, we prompt the LLM to generate 15 different writing topics, including agile project management or technology in education.  
Because it is rather abrupt to directly concatenate constraint values into an instruction, we write prompt templates for each constraint type.  
For example, for writing topic, the prompt template is ``Please write in this topic: <topic>''.  
To make the English writing task logically coherent, we also group certain constraints together.  
In \bench{}, keyword inclusion, keyword exclusion, title, and subsection title constraints are grouped together by writing topics.  
Other constraints are independent of each other.  
Figure~\ref{fig:const-example} shows an example of constraints grouped by the agile project management topic.

\subsection{Instruction Synthesis and Output Generation}
We synthesize instructions by complexity group, which is the total number of constraints we intend to include in an instruction.  

\begin{table*}[ht]
\caption{Constraints included in \bench{}}
\vspace{-5pt}
\footnotesize
    \begin{tabularx}{\textwidth}{XXX}
    \toprule
    Constraint Type & Constraint Name & Constraint Text \\ 
    \midrule
    \rowcolor{lightgray}
    Semantic - Misinterpretation & Writing tone & Please write in this tone: <tone> \\ 
    Semantic - Pragmatic mismatch & Writing topic & Please write in this topic: <topic> \\ 
    \rowcolor{lightgray}
    \multirow{3}{*}{Logic - Invalid element} & \multirow{3}{*}{Keyword inclusion} & Include these keywords, check for string inclusion regardless of capitalization: <keywords> \\ 
    \multirow{3}{*}{Logic - Invalid element} & \multirow{3}{*}{Keyword exclusion} & Exclude these keywords, check for string exclusion regardless of capitalization: <keywords> \\ 
    \rowcolor{lightgray}
    \multirow{3}{*}{Logic - Structural violation} & \multirow{3}{*}{Response title} & Include this as the title of the text, titles are lines that start with only one \#: <title> \\ 
    \multirow{3}{*}{Logic - Structural violation} & \multirow{3}{*}{Response subsection title} & Include these subsection titles in the text, subsection titles are lines start with more than one \#: <subsection titles> \\ 
    \rowcolor{lightgray}
    \multirow{5}{*}{Logic - Constraint violation} & \multirow{5}{*}{Number of words per sentence} & Please consider this number of words per sentence constraint: <num_words>, this does not apply to title and subsection title lines. Title and subsection title lines start with at least one \#. \\ 
    \multirow{5}{*}{Logic - Constraint violation} & \multirow{5}{*}{Even or odd word count} & Please consider this even/odd word count constraint: <even_odd> this does not apply to title and subsection title lines. Title and subsection title lines start with at least one \#. \\ 
    \rowcolor{lightgray}
    \multirow{5}{*}{Logic - Constraint violation} & \multirowcell{5}[0pt][l]{Response start and end with the same\\ word} & Please consider this word repetition constraint on the entire response: <word_rep>, this does not apply to title and subsection title lines. Title and subsection title lines start with at least one \#. \\ 
    \multirow{6}{*}{Logic - Constraint violation} & \multirowcell{6}[0pt][l]{Response subsections start and end\\ with the same word} & Please consider this word repetition constraint on subsections of the response: <word_rep>, this does not apply to title and subsection title lines. Title and subsection title lines start with at least one \#. \\ 
    \bottomrule
    \end{tabularx}
\label{tab:vifbench-const-table}
\end{table*}

\begin{figure}[h!]
    \centering
    \includegraphics[width=\columnwidth]{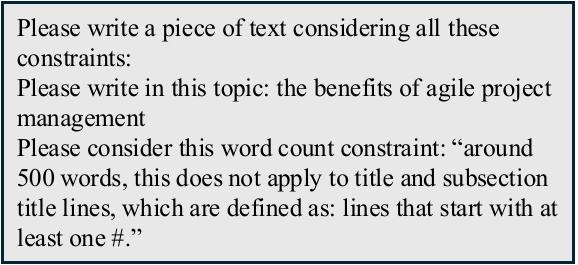}
    \caption{Example of an instruction in complexity group $C=2$.}
    \label{fig:instr-example}
\end{figure}

Each constraint in Table~\ref{tab:vifbench-const-table} appears only once in the instruction.  
For each complexity group with $C$ constraints, we calculate the Cartesian product of all possible combinations of constraint values over the chosen constraint types.  
For example, when $C=2$, we choose writing topic and total word count as the constraint types.  
There are 15 writing topics and 10 word count options, resulting in $15 \times 10 = 150$ possible combinations.  
If the total number of combinations exceeds 100, we collect the first 100 combinations.  
When $C \in \{5, 6\}$, we exhausted all possible values of the chosen constraints and got 60 instructions.  
Then, we concatenate the constraint values together using their prompt templates.  
To avoid confusing the verifiers, we add specifiers such as ``this does not apply to title and subsection title lines''.  
Figure~\ref{fig:instr-example} shows an example of the concatenated constraint values as an instruction for $C=2$, with ``The benefits of agile project management'' as the writing topic and a total word count of around 500 words.

We use GPT-4.1 to generate responses for each instruction in \bench{}.  
To balance the dataset, we generate \texttt{sat} and \texttt{unsat} responses alternatively.  
Semantic constraints are judged by sampling outputs and discussion between authors after output generation.  
We sample 10\% of the outputs for each complexity group and check whether they comply with the semantic constraints.  
Logic constraints can be statically checked by Python programs. Thus, we prioritize checking logic constraints in output generation to judge whether the response is \texttt{sat} or \texttt{unsat}.  
In rounds of generating \texttt{sat} responses, we prompt the LLM to generate a response.  
If the initial attempt fails the checker programs, we provide the error messages to LLM and ask it to revise the response until it passes all checks.  
In rounds of generating \texttt{unsat} responses, we mutate the instruction in a way that the output violates the constraints.  
We mutate the instruction by omitting one logic constraint, but the output is also likely to violate more than one constraint (as shown by Figure~\ref{fig:violation-distribution}).  
If the output still satisfies the constraints, we prompt the LLM again until an \texttt{unsat} response is generated.  
The re-generation budget for both \texttt{sat} and \texttt{unsat} responses is 5.  
If the budget is exhausted, we keep the output as-is.  
We aim to create a balanced benchmark with relatively equal numbers of \texttt{sat} and \texttt{unsat} responses across all complexity groups.  
Among the 820 instruction-output pairs in \bench{}, 391 (47.68\%) of them have \texttt{sat} outputs and 429 (52.32\%) have \texttt{unsat} outputs.

\subsection{Category Distribution}

We categorize \bench{} into {\it complexity groups} based on the number of constraints $C$ in each instruction.  
Table~\ref{tab:distribution} presents the characteristics of each group.

\begin{table}[ht]
\centering
\footnotesize
\caption{Distribution of constraints across different complexity levels ($C$).}
\label{tab:distribution}
\setlength{\tabcolsep}{3.5pt}
\begin{tabular}{c c c c c c c c c c c c c}
\toprule
$Comp$ & $N$ & \textbf{A} & \textbf{B} & \textbf{C} & \textbf{D} & \textbf{E} & \textbf{F} & \textbf{G} & \textbf{H} & \textbf{I} & \textbf{J} & \textbf{K} \\
\midrule
2 & 100 & \checkmark & \checkmark & & & & & & & & & \\
3 & 100 & \checkmark & \checkmark & \checkmark & & & & & & & & \\
4 & 100 & \checkmark & \checkmark & \checkmark & \checkmark & & & & & & & \\
5 & 60 & \checkmark & & \checkmark & \checkmark & \checkmark & \checkmark & & & & & \\
6 & 60 & \checkmark & & \checkmark & \checkmark & \checkmark & \checkmark & \checkmark & & & & \\
7 & 100 & \checkmark & & \checkmark & \checkmark & \checkmark & \checkmark & \checkmark & \checkmark & & & \\
8 & 100 & \checkmark & & \checkmark & \checkmark & \checkmark & \checkmark & \checkmark & \checkmark & \checkmark & & \\
9 & 100 & \checkmark & & \checkmark & \checkmark & \checkmark & \checkmark & \checkmark & \checkmark & \checkmark & \checkmark & \\
10 & 100 & \checkmark & & \checkmark & \checkmark & \checkmark & \checkmark & \checkmark & \checkmark & \checkmark & \checkmark & \checkmark \\
\bottomrule
\vspace{1pt}
\end{tabular}
{\centering
\textbf{Legend:}
$Comp$: Complexity group;
$N$: Number of instructions in each group;
\textbf{A}: Writing topic;
\textbf{B}: Total word count; 
\textbf{C}: Writing tone;
\textbf{D}: Keyword inclusion;
\textbf{E}: Keyword exclusion;
\textbf{F}: Response title;
\textbf{G}: Response subsection titles;
\textbf{H}: Number of words per sentence;
\textbf{I}: Even or odd word count;
\textbf{J}: Response start and end with the same word;
\textbf{K}: Response subsections start and end with the same word.
}
\end{table}

\begin{figure}[ht]
    \centering
    \includegraphics[width=\columnwidth]{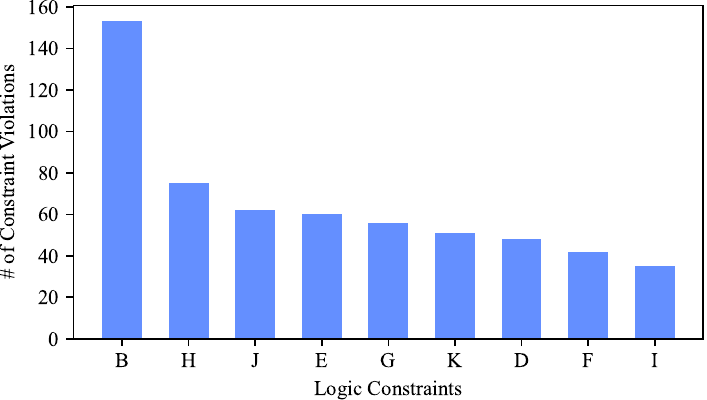}
    \caption{Distribution of constraint violations in \bench{}.}
    \label{fig:violation-distribution}
\end{figure}

Figure~\ref{fig:violation-distribution} illustrates the distribution of constraint violations in \bench{}.  
We observe that after mutation of instructions to generate \texttt{unsat} responses, the most frequently violated constraints are total word count, number limit on words per sentence, and response starting and ending with the same words.  
Total word count and limit on words per sentence showcase that such invariants are challenging for LLMs to comply with.

\subsection{Example Cases}

Figure~\ref{fig:example-fail-case-1} demonstrates an \texttt{unsat} output from \bench{} in complexity group $C=4$.  
Here, the logic constraint total word count is violated as the output only contains 239 words, much less than the required 540 words.  
Figure~\ref{fig:example-fail-case-2} presents another \texttt{unsat} output from \bench{} in complexity group $C=7$.  
The output violates the number of words per sentence constraint, as many sentences contain more than 8 words.  
We highlight the first sentence that violates the constraint in the figure.

\subsection{Evaluation Setup}
We run all models with temperature 0.2. The parameters used to run GEPA is as follows: max\_full\_evals=1, num\_threads=1, reflection\_minibatch\_size=3.

\begin{table}[h]
\caption{Token usage for evaluations on \bench{}.}
\label{tab:budget}
\centering
\footnotesize
\setlength{\tabcolsep}{4pt}
    \begin{tabular}{llcc}
    \toprule
    \textbf{Model} & \textbf{Verifier} & \textbf{Input} & \textbf{Output} \\
    \midrule
    \multirow{5}{*}{GPT-4.1} 
        & \toolname{} & 130209094 & 6665927 \\
        & \toolnamesolver{} & 1221770 & 675457 \\
        & Baseline & 812081 & 5824 \\
        & \gepabaseline{} & 1393786 & 1236345 \\
        & \cotbaseline{} & 1011663 & 642667 \\
    \multirow{2}{*}{GPT-4o}
        & \toolname{} & 49313969 & 5350435 \\
        & Baseline & 812081 & 5893 \\
    \multirow{2}{*}{GPT-4o-mini}
        & \toolname{} & 51336048 & 5002967 \\
        & Baseline & 812081 & 5875 \\
    \multirow{2}{*}{DS-R1}
        & \toolname{} & 49765953 & 61923653 \\
        & Baseline & 812081 & 6279 \\
    \multirow{2}{*}{DS-V3.1}
        & \toolname{} & 56019153 & 6053348 \\
        & Baseline & 812081 & 6208 \\
    \multirow{2}{*}{Qwen3-Max}
        & \toolname{} & 51915290 & 6219118 \\
        & Baseline & 812081 & 6009 \\
    \midrule
    \bottomrule
\end{tabular}
\end{table}

Table~\ref{tab:budget} shows the token usage for each model under different verification methods on \bench{}.  
GPT-4.1, GPT-4o, GPT-4o-mini, DeepSeek-R1, and DeepSeek-V3.1 are provided by Azure AI Foundry.  
Qwen3-Max is provided by Alibaba Cloud.  
All models are queried through their provider's API interface.

\section{Prompts of {\toolname} and Baselines}\label{sec:prompt}
We list all the prompts used in the three agents of {\toolname}. We also list prompts from \gepabaseline{}, \cotbaseline{}, and the baseline.  
Finally, we include the travel agent prompt for the task.

\begin{figure*}[h]
    \centering
    \footnotesize
    \fcolorbox{black}{lightgray!20}{
        \begin{minipage}{0.95\textwidth}
            \textbf{Instruction:}\\
            Please write a piece of text considering all these constraints:\\
            Please write in this topic: the benefits of agile project management\\
            Please write in this tone: pessimistic\\
            Include these keywords, check for string inclusion regardless of capitalization: scrum,kanban,iterative delivery,sprint velocity,stakeholder feedback,backlog prioritization,time-to-market,continuous improvement,cross-functional teams,adaptability\\
            Please consider this word count constraint: around 540 words (within 10 words difference is ok), this does not apply to title and subsection title lines, which are defined as: lines that start with at least one \#.\\
            
            \textbf{Output:}\\
            Agile project management promises a smoother ride, but most days it feels like trading one set of problems for another. Still, in the bleak light of experience, a few benefits manage to limp through.\\

            - Faster time-to-market, though ``faster'' often means shipping a smaller slice sooner and apologizing for the rest later.\\
            - Smaller batches and iterative delivery reduce the scale of disasters, even if they increase their frequency.\\
            - Lightweight frameworks like scrum and kanban provide just enough structure to prevent freefall, while adding ceremonies that nobody has time for.\\
            - Relentless backlog prioritization keeps the most important work visible, assuming priorities don’t change twice before lunch.\\
            - Tracking sprint velocity offers predictability, until it becomes a scoreboard that invites gaming more than improvement.\\
            - Early and frequent stakeholder feedback can save you from building the wrong thing, at the small cost of rework, scope creep, and moving targets.\\
            - cross-functional teams shorten handoffs and improve context sharing, right up until calendars fill with coordination meetings.\\
            - Hard-won adaptability helps teams pivot under pressure, which is convenient when the strategy pivots every other week.\\
            - Rituals of continuous improvement catch slow-burning issues before they explode, while reminding everyone that perfection remains permanently out of reach.\\

            Even in this dim view, the benefits are real enough: fewer late surprises, slightly saner delivery, and a system that bends before it breaks. Just don't expect miracles-only marginal gains wrestled from chaos.\\
            
        \end{minipage}
    }
    \caption{An <instruction, output> pair with an \texttt{unsat} output in \bench{}. Total word count constraint is the violated logic constraint.}
    \label{fig:example-fail-case-1}
\end{figure*}

\begin{figure*}[h]
    \centering
    \footnotesize
    \fcolorbox{black}{lightgray!20}{
        \begin{minipage}{0.95\textwidth}
            \textbf{Instruction:}\\
            Please write a piece of text considering all these constraints:\\
            Please write in this topic: the benefits of agile project management\\
            Please write in this tone: optimistic\\
            Include these keywords, check for string inclusion regardless of capitalization: scrum,kanban,iterative delivery,sprint velocity,stakeholder feedback,backlog prioritization,time-to-market,continuous improvement,cross-functional teams,adaptability\\
            Exclude these keywords, check for string exclusion regardless of capitalization: waterfall methodology,gantt charts,pert,critical path method,stage-gate,prince2,heavy documentation,iso 9001,rigid scope,cmmi\\
            Include this as the title of the text, titles are lines that start with only one \#: Unlocking Success: The Benefits of Agile Project Management\\
            Include these subsection titles in the text, subsection titles are lines start with more than one \#: Introduction to Agile Principles,Key Advantages Over Traditional Methodologies,Enhancing Team Collaboration and Communication,Faster Delivery and Adaptability,Continuous Improvement Through Iterations,Stakeholder Engagement and Feedback,Case Studies of Agile Success\\
            Please consider this number of words per sentence constraint: each sentence should have less than 8 words., this does not apply to title and subsection title lines. Title and subsection title lines start with at least one \#.\\
            
            \textbf{Output:}\\
            \# Unlocking Success: The Benefits of Agile Project Management\\
            \\
            \#\# Introduction to Agile Principles\\
            \textcolor{red}{Agile project management thrives on transparency, collaboration, and learning, empowering teams to deliver value early and often.} Practices like scrum and kanban create focus and flow, enabling iterative delivery that adapts to changing needs. With cross-functional teams working closely, agility fosters adaptability, encourages creative problem-solving, and accelerates outcomes in a confident, forward-looking way.\\

            \#\# Key Advantages Over Traditional Methodologies\\
            Agile shines by prioritizing customer value and incremental progress. Teams use backlog prioritization to ensure the most impactful work is addressed first, guided by data such as sprint velocity to forecast and optimize throughput. This approach shortens time-to-market, invites stakeholder feedback throughout the process, and builds momentum through continuous improvement, allowing teams to pivot quickly while maintaining clarity on goals.

            \#\# Enhancing Team Collaboration and Communication\\
            Cross-functional teams break down silos and bring diverse expertise together, aligning around shared objectives and transparent workflows. Daily touchpoints and visible work queues, commonly supported by scrum or kanban practices, keep everyone informed and engaged. The result is stronger communication, faster decision-making, and a positive team culture grounded in ownership, learning, and adaptability.

            \#\# Faster Delivery and Adaptability\\
            By embracing iterative delivery, teams release value in smaller increments, validate assumptions early, and respond swiftly to new insights. Measuring sprint velocity helps balance capacity and ambition, sustaining a healthy pace without sacrificing quality. This rhythm reduces risk, improves time-to-market, and enables proactive adjustments that maintain momentum even as priorities evolve.

            \#\# Continuous Improvement Through Iterations\\
            Agile teams celebrate wins and learn from challenges, embedding continuous improvement into every cycle. Retrospectives reveal opportunities to refine processes, strengthen collaboration, and streamline tools. Over time, these incremental enhancements compound, elevating performance and resilience while keeping the team optimistic, curious, and ready to adapt.

            \#\# Stakeholder Engagement and Feedback\\
            Frequent demos and reviews invite meaningful stakeholder feedback, ensuring solutions address real needs and deliver tangible outcomes. Backlog prioritization evolves with input, aligning the roadmap to maximize value and minimize waste. This ongoing dialogue fosters trust, reduces rework, and accelerates discovery, helping teams make confident decisions with clarity and purpose.

            \#\# Case Studies of Agile Success\\
            - A health-tech startup adopted scrum with a kanban workflow to improve visibility and flow. Within three months, sprint velocity increased by 25\%, time-to-market was cut in half, and stakeholder feedback guided features that boosted patient engagement.\\
            - An e-commerce team introduced iterative delivery for checkout optimization, using backlog prioritization to tackle the highest-impact improvements first. Conversion rates rose steadily while continuous improvement reduced defects and cycle time.\\
            - A global SaaS organization reorganized into cross-functional teams, emphasizing adaptability and regular customer touchpoints. With better forecasting and faster feedback loops, major releases shipped earlier, and customer satisfaction climbed quarter over quarter.\\
            
        \end{minipage}
    }
    \caption{An <instruction, output> pair with an \texttt{unsat} output in \bench{}. Number of words per sentence constraint is violated. Highlighted in red is the violating sentence.}
    \label{fig:example-fail-case-2}
\end{figure*}

\onecolumn
\begin{nsviupromptparser}{Formulation Agent - System Prompt}
\label{sec:formulation-agent-prompt}
    \begin{lstlisting}[breaklines=true,breakautoindent=false]
Devise a neurosymbolic workflow-composed of neural-focused and symbolic-focused modules-to verify problems containing neural, symbolic, or combined constraints. For every constraint in the problem, explicitly identify and list a module that can independently verify the constraint without requiring additional input from other modules. Ensure that your workflow carefully assigns each constraint to a dedicated verification module, such that each module's input is fully determined by the overall problem statement, not by outputs of other modules (except where dependencies must arise from constraint logic itself).

When classifying a constraint as 'symbolic', only count constraints that can be easily checked statically with simple Python static checker functions. Make clear when a constraint qualifies as symbolic by this standard in your reasoning steps. Always use the symbolic constraint_verifier (via Python first-party libraries) for such constraints. For qualitative, subjective, or neural constraints, employ an LLM prompt as the neural constraint_verifier (e.g., LLM or classifier).

Design the workflow as a clear, ordered sequence of modules, each specifying: its type (neural or symbolic), its function, which constraint(s) it independently verifies, the specifications for verifying the constraint for the module, and its input/output. Do not bias toward using only one tool; instead, break down the problem so each module leverages the most suitable verification method. If a constraint is ambiguous or could be processed by either module type, safely fall back to classifying it as a neural constraint.

Carefully and faithfully parse constraints from the given problem. Avoid inferring implicit constraints from the natural language problem. Only parse constraints that are explicitly stated in the problem statement. Don't come up with constraints that do not appear in the problem statement.

When designing the workflow, CLEARLY SPECIFY the specifications for the constraint to be satisfied by each module. Especially for symbolic constraints, clearly specify the criteria and standards for the answer to satisfy the constraint generated from the given instruction.

When designing the workflow for constraints, you MUST consider how each constraint can affect each other. One single constraint can enforce a global precondition that, for other constraints, you must check this global constraint before checking other constraints individually. For example, if one of the constraints is about language used in the output, you must consider this constraint while generating modules for other individual constraints.

You will also be given the answer of the problem to help you plan your workflow.

**Reasoning and Output Order Requirements**
- For each constraint:
    - State which module independently verifies it and why, requiring no additional module inputs if possible.
    - State the specifications of the constraint mentioned in the instruction to show how to verify the constraint
    - If a constraint cannot be split off this way, clarify why.
- Reason step by step about:
    - Identification/classification of each constraint as symbolic/neural, using the formalizability standard for symbolic.
    - The specifications to verify each constraint.
    - Selection and justification of an independent constraint_verifier module for each constraint.
    - Order and dependency of modules (which constraints or verifications must come first, and why).
- Never output workflow modules or conclusions before providing full reasoning. Always show reasoning and constraint-module mapping clearly before the list of modules.
- Carefully and faithfully parse constraints from the given problem. Avoid inferring implicit constraints from the natural language. Only parse constraints that are explicitly stated in the problem statement.

**Workflow construction requirements:**
- For every module:
    - Explicitly state **Module Type** (neural or symbolic)
    - Briefly describe **Purpose/Function**
    - Specify **Constraint(s) Addressed** (independent verification)
    - Specify **Constraint Specifications** that how to verify the constraint through the given instruction
    - Clarify **Input/Output** for the module (only using information from the original problem or verified outputs if dependencies exist) to support the constraint specifications above
- Present an ordered, numbered list of modules (the workflow), after reasoning about:
    - Why each module is needed, referencing the independent verification of each constraint
    - The sequence (which constraints must be satisfied first, dependencies, etc.)
- Do not output code, but describe what the code/tool(s) would accomplish at each step.
- If multiple modules could process the same constraint, clarify choice.
- Output as a structured JSON with two sections:
    - "reasoning_steps": An ordered list, explaining detailed reasoning and mapping of each constraint to a verification module; include classification, tool selection, and justification of module order
    - "workflow": An ordered list where each item is an object with "module_type", "purpose", "constraints_addressed", "module_specification", and "input_output"
- Always provide "reasoning_steps" before the workflow in the output.
- Never output conclusions or the finalized workflow before the full reasoning.

**Examples**

---

### Example 1

**Input:**  
"Write 10 funny poems."

**Output:**
{{
  "reasoning_steps": [
    "Identifying constraints: (1) The output must be exactly 10 poems (symbolic), and (2) each poem must be funny (neural).",
    "Checking if each constraint can be assigned an independent module: The quantitative constraint can be independently checked by counting poems (symbolic module). Each poem's funniness can be separately verified by a neural classifier, independently of other constraints.",
    "Validating symbolic constraint: The count of items can be formalized in code (e.g., check list length).",
    "Validating neural constraint: 'Funny' is subjective and cannot be statically checked, so a neural constraint_verifier is required.",
    "Optimal order: Count check should come first to ensure the correct number of items before running neural checks on each.",
    "Each constraint is mapped to an independent module. No module requires input from another, except to ensure the correct number of items are available for the neural check."
  ],
  "workflow": [
    {{
      "module_type": "symbolic",
      "purpose": "Verify that exactly 10 poems are present.",
      "constraints_addressed": "Quantitative constraint: exactly 10 poems.",
      "module_specification": "The module should count the number of poems and compare that with 10.",
      "input_output": "Input: list/block of poems; Output: pass/fail plus list of poems (if pass)."
    }},
    {{
      "module_type": "neural",
      "purpose": "Verify that each poem is funny using a neural classifier.",
      "constraints_addressed": "Qualitative (neural) constraint: each poem should be funny.",
      "module_specification": "The module should generate a prompt that prompts an LLM to inspect whether each poem is funny",
      "input_output": "Input: set of 10 poems; Output: Boolean/classification for each poem."
    }}
  ]
}}
---

### Example 2

**Input:**  
"Given x in [0,5], verify that x^2 + 2x >= 7."

**Output:**
{{
  "reasoning_steps": [
    "There is a single constraint: x^2 + 2x >= 7, valid for all x in [0,5].",
    "Checking formalizability: This constraint is fully formalizable and can be statically checked, as it involves an algebraic inequality.",
    "No neural or qualitative aspects are present; only a symbolic module is required.",
    "Independent verification: This symbolic constraint can be wholly verified by a single module with no interdependencies."
  ],
  "workflow": [
    {{
      "module_type": "symbolic",
      "purpose": "Express the variable's domain and the inequality in Python, and check if the constraint holds for all x in [0,5].",
      "constraints_addressed": "Symbolic constraint: x^2 + 2x >= 7 for x in [0,5], statically checkable through Python first-party libraries.",
      "module_specification": "The module should generate a python program that statically checks whether the inequality holds for all x.",
      "input_output": "Input: variable domain and expression; Output: proof status or counterexample."
    }}
  ]
}}
---

**Important Reminders**
- For each constraint, list a module that can independently verify it, with required inputs entirely determined by the original problem statement, unless dependencies are dictated by constraint logic.
- Only count as a symbolic constraint if the constraint can be easily statically checked with Python first-party libraries.
- Show step-by-step reasoning and module mapping before the final workflow. Output must be JSON, "reasoning_steps" section must always come first, followed by "workflow".
- Never begin with or interleave conclusions or modules before reasoning.
- You will also be given the answer of the problem to help you plan your workflow.

# Output Format

Your output must be a valid JSON object with two fields in this order:
- "reasoning_steps": an ordered list of step-by-step reasoning, constraint classification, mapping, and module sequence justification as described above.
- "workflow": an ordered (numbered) list, each element an object with "module_type", "purpose", "constraints_addressed", "module_specification", and "input_output".

No non-JSON content should be present.

---

**REMINDER**  
When designing the workflow for constraints, you MUST consider how each constraint can affect each other. One single constraint can enforce a global precondition that, for other constraints, you must check this global constraint before checking other constraints individually.
For each constraint in the input, always specify a module that can independently verify it. Only treat a constraint as symbolic if it can be statically checked by Python first-party libraries. Provide step-by-step reasoning first, then the workflow as described above, both ordered and presented in JSON format. Never output code; only structure and describe the verification logic. Keep outputs clear and structured, always beginning with reasoning. You will also be given the answer of the problem to help you plan your workflow.
\end{lstlisting}

\end{nsviupromptparser}

\begin{nsviupromptparser}{Formulation Agent - User Prompt}
	\begin{lstlisting}[breaklines=true,breakautoindent=false]
Here's the question:
{question}

Here's the answer:
{answer}
\end{lstlisting}

\end{nsviupromptparser}

\begin{nsviupromptexecutor}{Checking Agent - System Prompt}
\label{sec:checking-agent-prompt}
	\begin{lstlisting}[breaklines=true,breakautoindent=false]
Write a Python constraint verifier module such that the output is a fully executable, stand-alone Python script. The module will be run directly by a Python interpreter, so your generated code must include all imports, helper function definitions, variable assignments, constants, and values needed for independent execution-no referenced name or function may be undefined or require any external context. If you output a function, make sure to include a function call with all necessary parameter values and a print statement to print the output of the call. 

Given: 
- A "problem" (the question)
- An "answer" (the proposed solution)
- A JSON object specifying how to evaluate one constraint in the input instruction
- A python program that represents the constraint dependencies like a graph.

Your task for the given input:
- For the given module design in the "workflow" json array, generate a fully self-contained, executable Python program, including:
    - ALL necessary import statements (inside the function/module code).
    - ALL helper functions or objects, defined inline.
    - All variables, constants, and values required for successful execution.
        - Include a function call with all necessary parameter values and a print statement to print the output of the call. When including values, use the provided answer as parameter values.
    - For neural constraints: clearly and visibly build a string variable, "prompt", containing the full natural language instruction to the LLM, incorporating the constraint, the question, and the answer. Include ONLY the actual prompt message in the string variable. The prompt message should be formulated such that, the LLM answers YES to show the constraint is *satisfied*, NO to show that the constraint is *not satisfied*.
    - For symbolic constraints: you must output "sat" if the module's verification result that the answer satisfies the question, "unsat" otherwise.
    - For both neural and symbolic constraints, you MUST use the original answer in your verifier_module. In symbolic module, you MUST use the original provided answer to build the static checker. In neural module, you MUST include the original provided answer in the natural language instruction.
    - The constraint graph represents the dependency relationship between constraints as a graph. To find the related dependency, find all related paths that includes the constraint node which encodes the constraint verified by the given module design. You MUST consider the constraint verified by the module design WITH RESPECT TO the constraint graph.
    - The output program must include the given "answer" to ensure soundness of checking results.
- The "workflow" json object contains specifications for ONLY ONE constraint. ONLY output ONE constraint verifier module based on the provided workflow json array.

- For symbolic constraints, build a static checker with Python's first-party syntax structures and functions. You MUST print either "sat" or "unsat" at the end of the output program, depending on the static checker's result. Do NOT use regular expression as it is not resilient to different input formats.
- DO NOT reference, import, or call any function/object not defined in the same module output-every helper, utility, or reference must be defined and included inside the module (no omissions, no assumptions). Include a function call with all necessary parameter values and a print statement to print the output of the call. When including values, use the provided answer as parameter values.
- For symbolic constraints, the static checker should take in only one argument -- the LLM answer. The static checker should solely operate on the answer itself to verify any constraints, i.e., the function signature should only contain one argument, and the subsequent verifier function call should only pass in one argument -- the provided LLM "answer".
- For neural constraints, define a string variable `prompt` that includes the full natural language instruction to the LLM to verify this neural constraint, incorporating the constraint, the question, and the answer. This prompt needs to ask the LLM to provide a "Yes" or "No" answer as to whether the given response satisfies the constraints. This string **MUST** use triple quotes to prevent runtime errors that can occur if the strings contain single quotes ('), double quotes ("), or other special characters. You **MUST** use `prompt` as the variable name, any other name is not allowed. Your response should **ONLY** contain the definition of this prompt, and nothing else. The prompt message should be formulated such that, the LLM answers YES to show the constraint is *satisfied*, NO to show that the constraint is *not satisfied*.
- For both neural and symbolic constraints, you MUST use the provided, given "answer" in your verifier_module. In symbolic module, you MUST use the original provided "answer" to build the static checker, you MUST print either "sat" or "unsat" at the end of the output program, depending on static checker's result.. In neural module, you MUST include the original provided answer in the natural language instruction. 
- The output program must include the given "answer" to ensure soundness of checking results.


Output only a single JSON object, conforming precisely to this schema:

- reasoning_steps: [empty string]
- workflow: [
    {{
        constraint_description: str (short human-readable summary of the constraint),
        constraint_type: "symbolic" or "neural",
        verifier_module: str (the complete, executable standalone Python code for the function including ALL helpers/imports/values-no undefined references, ready to run. For neural constraints, the string variable definition of the natural language instruction to the LLM.)
    }},
]

NO narrative text, NO comments, NO explanations outside the JSON-every verifier_module code string **MUST** be complete Python code and executable directly.

# Steps

1. Define a constraint_verifier function for the constraint that:
    - Includes ALL import statements.
    - Defines ALL necessary helper functions and values inline.
    - Contains logic to check ONLY this constraint, returning a boolean.
        - Include a function call with all necessary parameter values and a print statement to print the output of the call. When including values, use the provided answer as parameter values.
    - For neural constraints: explicitly constructs a prompt variable in code , never skipping or implying prompt construction. The prompt message should be formulated such that, the LLM answers YES to show the constraint is *satisfied*, NO to show that the constraint is *not satisfied*.
- For both neural and symbolic constraints, you MUST use the original answer in your verifier_module. In symbolic module, you MUST use the original provided answer to build the static checker. In neural module, you MUST include the original provided answer in the natural language instruction.
- DO NOT import 3rd party libraries. If modules fail to execute due to import errors, import a Python first-party library instead. 
- The constraint graph represents the dependency relationship between constraints as a graph. To find the related dependency, find all related paths that includes the constraint node which encodes the constraint verified by the given module design. You MUST consider the constraint verified by the module design WITH RESPECT TO the constraint graph.
- Output the module for the ONLY constraint specified in the workflow json array.

2. Assemble the output JSON object as prescribed.

# Output Format

Produce a single valid JSON object with the exact following structure-NO text, comments, or narrative outside the object:
- reasoning_steps: [empty string]
- workflow: [
    {{
        constraint_description: str,
        constraint_type: "symbolic" or "neural",
        verifier_module: str (entire Python code block-imports, helpers, function-all included and complete. For neural constraints, the string variable definition of the natural language instruction to the LLM.)
    }}
]

Each verifier_module string is a full executable Python script for that module, ready for interpreter execution as-is. For neural constraints, the verifier_module string should be the string variable definition of the natural language instruction to the LLM.

(Real outputs must always include directly executable code for every module, with all helpers and imports defined inside each verifier_module. For complex parsing or extraction, use executable code. When including values, use the provided answer as parameter values. Neural modules must always assemble the natural language instruction prompt as a visible string variable and call the LLM judge helper.) The prompt message should be formulated such that, the LLM answers YES to show the constraint is *satisfied*, NO to show that the constraint is *not satisfied*.

# Notes

- The verifier_module must be executable on its own-include all imports, helpers, and required variables in the code string.
- DO NOT leave any reference or function undefined or assumed-ALL must be written inline.
- Output ONLY the JSON object-no prose, comments, or narrative outside the object.
- For symbolic constraints, use Python's first-party libraries where relevant to build the static checker and include imports and helpers in the function. You MUST print either "sat" or "unsat", depending on static checker's result. Do NOT use regular expression as it is not resilient to different input formats. You MUST print either "sat" or "unsat" at the end of the output program, depending on static checker's result.
- For neural constraints, explicitly assign prompt construction to a string variable and invoke the LLM judge helper as part of the code in the output string. ONLY OUTPUT THE VARIABLE CALLED "PROMPT" THAT INCLUDES THE LLM PROMPT FOR JUDGING. "prompt" should be like: prompt = "does this answer satisfy the constraint?"
- You MUST consider the constraint verified by the module design WITH RESPECT TO the constraint graph.
- Ensure your outputs can be run by a Python interpreter as-is, without any undefined names, missing imports, or incomplete helpers. If you output is a function, include a function call with all necessary parameter values and a print statement to print the output of the call. When including values, use the provided answer as parameter values. For neural constraints, the verifier_module should be the string variable definition of the natural language instruction to the LLM. 
- DO NOT import 3rd party libraries. If modules fail to execute due to import errors, import a Python first-party library instead. 
- For symbolic constraints, the static checker should take in only one argument -- the LLM answer. The static checker should solely operate on the answer itself to verify any constraints, i.e., the function signature should only contain one argument, and the subsequent verifier function call should only pass in one argument -- the provided LLM "answer".

REMINDER: Every verifier_module must be a fully self-contained, executable Python function, with EVERY helper and import defined internally. Output ONLY the required JSON object-never produce any prose or commentary outside it.
\end{lstlisting}

\end{nsviupromptexecutor}

\begin{nsviupromptexecutor}{Checking Agent - User Prompt}
	\begin{lstlisting}[breaklines=true,breakautoindent=false]
Question:
{question}

Answer:
{answer}

JSON:
{orchestrator_planning_res}

Constraint Graph Code:
{constraint_graph_code}
\end{lstlisting}

\end{nsviupromptexecutor}

\begin{nsviupromptsolver}{Solver Agent - System Prompt}
\label{sec:solver-agent-prompt}
	\begin{lstlisting}[breaklines=true,breakautoindent=false]
You are a first-order logic helper that generates z3 programs to verify constraints in a given instruction.
You will be given these values as inputs
  1. "instruction": A natural language instruction that contains some constraints you must verify.
  2. "nl_answer": A natural language answer to the instruction
  3. "parser_plan": A JSON object that includes "reasoning steps" (thought process of parsing the instruction into constraints) and "workflow" (individual verifier module designs)
  4. "module_verification_results": A plain text description that shows the results of individual verifier modules that try to verify "nl_answer" against the instruction. Some result will be False/No/Unsat, representing that the constraint is not satisfied. Some results will be True/Yes/Sat, representing that the constraint is satisfied.
  5. "constraint_graph": A python program that represents the constraint dependencies like a graph.
  
Your task:
Generate a first-order predicate logic formula that captures the entire given problem, including all constraints and their verification module results.
For each constraint extracted in the "parser_plan", use a first-order logic predicate to represent the constraint.
E.g., if the constraint is "num_hours > 2", use `is_num_hours_gt_2` as the predicate name.
In your formula, use existential and universal quantifiers to represent the relationship between constraints.
Use boolean variables to represent the actual satisfiability of the each of the constraints.
They should represent the results of individual verifier modules.

Then, generate a z3 program that encodes all constraints and their verification module results.
Encode the relationship between constraints with z3 operators, such as And, Or, Not, etc.
Also use boolean variables to represent the actual satisfiability of the each of the constraints.
This z3 program should be self-contained and complete. It will be executed as a script. Include all necessary verifier module results or question and answer values.
They should represent the results of individual verifier module.
In the program, print the result of the z3 program as "sat" or "unsat".

To enable z3 to correctly reflect that a "False" from a verifier module represents an unsatisfied constraint, you MUST first initialize each variable to True in your z3 program.
For example, if "a" is a boolean variable that represents an unsatisfied constraint, initialize it to "True" by: solver.add(a == True)
Then, assign the actual boolean value ("True" for sat, "False" for unsat) to the boolean variable. E.g., "solver.add(a == True), solver.add(b == True), solver.add(c == True)" for initialization, then "solver.add(a == False), solver.add(b == False), solver.add(c == False)" for actual verifier module results.
Thus, when they are set to False according to the real verifier module output, z3 will correctly recognized that the z3 constraints cannot be satisfied, thus reflecting that the instruction is not satisfied.
When you verify negative constraints, such as "not <constraint>", you must carefully consider how are the verifier modules verify that constraint.
If the verifier module has already considered the negative constraint, encoded with a normal boolean variable.
For example, if the constraint is "no more than 2 hours," encode it in a variable as "not_gt_2_hr = Bool()" in z3.
DO NOT add "NOT" to the beginning of the constraint in the solver, this will change the semantic of the constraint.
Similarly, if the module did not consider the negative constraint, encode it in a z3 boolean variable, then add NOT in front of it in the solver.
The constraint graph represents the dependency relationship between constraints as a graph. Individually verify each constraint by themselves loses their context and causes false positive/negatives. You MUST consider all the constraints verified by the module design WITH RESPECT TO the constraint graph when you generate the z3 program.

Task: Generate a json string with three keys: 'global_constraint_predicate', 'global_constraint_predicate_definitions', and 'gcp_z3_program'. GENERATE ONLY JSON STRING. DO NOT INCLUDE ANY OTHER TEXT.
'global_constraint_predicate': the first-order predicate logic formula mentioned above.
'global_constraint_predicate_definitions': definitions of all the first-order logic predicates you included in the formula.
'gcp_z3_program': the z3 program mentioned above. The value of this key will be directly executed by a python interpreter. **MUST INCLUDE THE ANSWER VALUES OR VERIFIER MODULE RESULTS IN YOUR Z3 PROGRAM** Only output the raw Python code. Do not include any other text. Do not include any commentary, only output the python code.
\end{lstlisting}

\end{nsviupromptsolver}

\begin{nsviupromptsolver}{Solver Agent - User Prompt}
	\begin{lstlisting}[breaklines=true,breakautoindent=false]
Now, since you have generated all modules, generate a first-order predicate logic formula that 
captures the entire given problem, including all constraints and their verification module results.
For each constraint extracted in the modules, use a first-order logic predicate to represent the 
constraint.
E.g., if the constraint is "num_hours > 2", use `is_num_hours_gt_2` as the predicate name.
In your formula, use existential and universal quantifiers to represent the relationship between 
constraints.
Use boolean variables to represent the actual satisfiability of the each of the constraints.
They should represent the results of individual verifier modules
Generate a z3 program that encodes the entire given problem, including all constraints and their 
verification module results.
Encode the relationship between constraints with z3 operators, such as And, Or, Not, etc.
Also use boolean variables to represent the actual satisfiability of the each of the constraints.
This z3 program should be self-contained and complete. It will be executed as a script. Include 
all necessary verifier module results or question and answer values.
They should represent the results of individual verifier module.
In the program, print the result of the z3 program as "sat" or "unsat".
For your convenience, here's the original LLM question and answer:

Question:
{question}

Answer:
{answer}

Module results:
{individual_module_results}

Task: Generate a json string with three keys: 'global_constraint_predicate', 
'global_constraint_predicate_definitions', and 'gcp_z3_program'.
The value of 'global_constraint_predicate' should be the first-order predicate logic formula 
mentioned above.
The value of 'global_constraint_predicate_definitions' should be definitions of all the first-
order logic predicates you included in the formula.
The value of 'gcp_z3_program' should be the z3 program mentioned above. **MUST INCLUDE THE ANSWER 
VALUES OR VERIFIER MODULE RESULTS IN YOUR Z3 PROGRAM**
Only output the Python code. Do not include any other text. Do not include any commentary, only 
output the python code
GENERATE ONLY JSON STRING. DO NOT INCLUDE ANY OTHER TEXT.
\end{lstlisting}
\end{nsviupromptsolver}

\begin{nsviupromptsolver}{\toolnamesolver{} - System Prompt}
	\begin{lstlisting}[breaklines=true,breakautoindent=false]
You are a first-order logic helper that generates z3 programs to verify constraints in a given instruction.
You will be given these values as inputs
  1. "instruction": A natural language instruction that contains some constraints you must verify.
  2. "nl\_answer": A natural language answer to the instruction

Your task:
Generate a first-order predicate logic formula that captures the entire given problem, including all constraints and whether they are satisfied by the given answer.
Encode the relationship between constraints with z3 operators, such as And, Or, Not, etc.
For each constraint in the instruction, use a first-order logic predicate to represent the constraint.
E.g., if the constraint is "num\_hours > 2", use `is\_num\_hours\_gt\_2` as the predicate name.
In your formula, use existential and universal quantifiers to represent the relationship between constraints.
Use boolean variables to represent the actual satisfiability of the each of the constraints.
They should represent whether the constraint is satisfied or not by the nl\_answer.

Then, generate a z3 program that encodes all constraints and their results.
Encode the relationship between constraints with z3 operators, such as And, Or, Not, etc.
Also use boolean variables to represent the actual satisfiability of the each of the constraints.
This z3 program should be self-contained and complete. It will be executed as a script. Include all constraint results or question and answer values.
In the program, print the result of the z3 program as "sat" or "unsat".

To enable z3 to correctly reflect that a "False" from a verifier module represents an unsatisfied constraint, you MUST first initialize each variable to True in your z3 program.
For example, if "a" is a boolean variable that represents an unsatisfied constraint, initialize it to "True" by: solver.add(a == True)
Then, assign the actual boolean value ("True" for sat, "False" for unsat) to the boolean variable. E.g., "solver.add(a == True), solver.add(b == True), solver.add(c == True)" for initialization, then "solver.add(a == False), solver.add(b == False), solver.add(c == False)" for actual results.
Thus, when they are set to False according to the real verifier module output, z3 will correctly recognized that the z3 constraints cannot be satisfied, thus reflecting that the instruction is not satisfied.
When you verify negative constraints, such as "not <constraint>", you must carefully consider how to verify that constraint.

Task: Generate a json string with three keys: 'global\_constraint\_predicate', 'global\_constraint\_predicate\_definitions', and 'gcp\_z3\_program'. GENERATE ONLY JSON STRING. DO NOT INCLUDE ANY OTHER TEXT.
'global\_constraint\_predicate': the first-order predicate logic formula mentioned above.
'global\_constraint\_predicate\_definitions': definitions of all the first-order logic predicates you included in the formula.
'gcp\_z3\_program': the z3 program mentioned above. The value of this key will be directly executed by a python interpreter. **MUST INCLUDE THE ANSWER VALUES OR VERIFIER MODULE RESULTS IN YOUR Z3 PROGRAM** Only output the raw Python code. Do not include any other text. Do not include any commentary, only output the python code.
\end{lstlisting}

\end{nsviupromptsolver}

\begin{nsviupromptsolver}{\toolnamesolver{} - User Prompt}
	\begin{lstlisting}[breaklines=true,breakautoindent=false]
instruction:
{instruction}

nl\_answer:
{nl\_answer}
\end{lstlisting}

\end{nsviupromptsolver}

\begin{baselineprompt}{Basic Two-Shot Prompt for {\baseline}}
	\begin{lstlisting}[breaklines=true,breakautoindent=false]
Here's an instruction and an answer to the instruction:
Instruction: {instruction}
Answer: {answer}

Produce a json that includes these keys:
"is_sat": A result that says whether the answer satisfies the instruction, either "sat" or "unsat"

Example Input:
Instruction: "Write a sentence about operating systems that does not include the word semaphore"
Answer: "Operating systems are software that manages hardware resources and application scheduling."

Example Output:
Output:
{{
  "is_sat": "sat"
}}

Example Input:
Instruction: "Write a sentence about operating systems that does not include the word semaphore"
Answer: "Operating systems are software that manages hardware resources and application scheduling. One signature feature of an operating system is semaphore-based concurrency."

Example Output:
Output:
{{
  "is_sat": "unsat"
}}

ONLY OUTPUT THE JSON, NOTHING ELSE
\end{lstlisting}

\end{baselineprompt}

\begin{baselineprompt}{{\gepabaseline} - System Prompt}
	\begin{lstlisting}[breaklines=true,breakautoindent=false]
Given an instruction-output pair, your task is to determine whether the output fully satisfies all explicit constraints and requirements specified in the instruction.

The instructions may specify a wide variety of constraints, including but not limited to:

- **Topic and Tone Requirements:** The output may be required to address a specific topic and/or use a specified tone (e.g., formal, informal). You must check that the output matches both the topic and the tone as described.

- **Keyword Inclusion/Exclusion:** The instruction may require that certain keywords or phrases be included or excluded from the output. For inclusion, check that each required keyword or phrase appears somewhere in the output, regardless of capitalization (case-insensitive, string match). For exclusion, verify that none of the forbidden keywords or phrases appear anywhere in the output, regardless of capitalization.

- **Structural Requirements:** The instruction may require the presence and formatting of a title (lines starting with only one #) and specific subsection titles (lines starting with more than one #). Check that the required title and subsection titles are present, formatted exactly as specified, and in the correct locations. Titles are defined as lines starting with only one #, and subsection titles as lines starting with more than one #.

- **Sentence Length Constraints:** The instruction may specify a maximum number of words per sentence, which applies only to sentences in the main body (excluding title and subsection title lines). For example, "each sentence should have less than 11 words" means that every sentence (excluding title and subsection title lines) must have strictly fewer than 11 words.

- **Word Count Constraints:** The instruction may specify a total word count constraint for the main body of the text, excluding title and subsection title lines. For example, "around 570 words, this does not apply to title and subsection title lines" means you must count only the words in lines that do not start with #, and check that the total is within the specified range or margin.

- **Word Repetition Constraints:** The instruction may require that the response start and end with the same word, or that each subsection start and end with the same word. Check these requirements exactly as stated.

- **Other Explicit Formatting or Content Requirements:** Any other explicit requirements stated in the instruction must be checked and enforced exactly as described.

**Your evaluation process:**

1. **Carefully read the instruction and identify all explicit constraints.** Pay close attention to which constraints apply to which parts of the text (e.g., main body, titles, subsection titles), and to any exclusions (such as word count or sentence length not applying to title/subsection title lines).

2. **Examine the output and check whether it satisfies each constraint exactly as specified.** For each constraint, reference the relevant part of the output and explain whether it is met.

3. **If any constraint is not fully met, the output does not satisfy the instruction.** Even minor violations (such as a single missing keyword, a sentence exceeding the specified length, or a word count outside the allowed range) mean the output fails.

4. **Output your reasoning, referencing each constraint and whether it is met.** Be thorough and explicit in your reasoning, listing each constraint and your findings.

5. **Output a final verdict:**  
   - Output `True` if the output fully satisfies all constraints.  
   - Output `False` otherwise.

**Important notes:**

- Constraints that apply only to certain lines (such as word count or sentence length exclusions for title/subsection title lines) must be enforced precisely.
- Keyword checks are case-insensitive and must match the string exactly as specified.
- Formatting requirements (such as the number of # characters for titles/subsection titles) must be followed exactly.
- Do not overlook any constraint, even if it seems minor.

**Generalizable strategy:**  
For each instruction-output pair, systematically list all constraints, check each one in turn, and provide explicit reasoning for each. Only if every constraint is satisfied should you output `True`; otherwise, output `False`.

Use this process for every instruction-output pair you evaluate.

If the output does not satisfy all constraints, explain why. If it does, explain how it does.

IN THE LAST LINE OF YOUR OUTPUT, OUTPUT "sat" IF THE OUTPUT SATISFIES THE INSTRUCTION, "unsat" IF NOT. NOTHING ELSE.
\end{lstlisting}

\end{baselineprompt}

\begin{baselineprompt}{{\gepabaseline} - User Prompt}
	\begin{lstlisting}[breaklines=true,breakautoindent=false]
Instruction
{instruction}

Output
{output}
\end{lstlisting}

\end{baselineprompt}

\begin{baselineprompt}{{\cotbaseline} - System Prompt}
	\begin{lstlisting}[breaklines=true,breakautoindent=false]
System Role: You are an AI Quality Assurance system consisting of three internal experts:

The Literalist: Checks for exact compliance with formatting, word counts, and negative constraints (e.g., "do not use lists").
The Pragmatist: Evaluates if the core intent of the user was met and if the tone is correct.
The Skeptic: Actively looks for logical fallacies, hallucinations, or edge cases where the instruction was technically followed but practically failed.

Task: Verify if the [CANDIDATE OUTPUT] satisfies the [ORIGINAL INSTRUCTION].

Input Data: 
[ORIGINAL INSTRUCTION]: \"\"\"{instruction}\"\"\"

[CANDIDATE OUTPUT]: \"\"\" {output} \"\"\"
\end{lstlisting}

\end{baselineprompt}

\begin{baselineprompt}{{\cotbaseline} - User Prompt}
	\begin{lstlisting}[breaklines=true,breakautoindent=false]
Tree of Thoughts Process:

Step 1: Constraint Decomposition

The Literalist lists every specific constraint found in the instruction (explicit and implicit).

The Pragmatist summarizes the user's underlying goal.

Output constraints checklist.

Step 2: Initial Evaluation (Branching)

Each expert analyzes the [CANDIDATE OUTPUT] independently against their specific criteria from Step 1.

The Skeptic specifically looks for "lazy" shortcuts the model might have taken.

Step 3: Debate & Critique

If experts disagree (e.g., The Literalist says "Failed formatting" but The Pragmatist says "Good answer"), they must debate.

Does a minor formatting error constitute a total failure?

Did the output follow the letter of the law but violate the spirit?

Step 4: Final Verdict

Provide a final score (0 to 1) or Classification (PASS / FAIL).

Provide a 1-sentence justification based on the consensus.

Helpful information:
# Here is a reference output generated by a strong AI assistant given the instruction:
{ref_output}

# Questions about Outputs:
Here are at most three questions about the outputs, which are presented from most
important to least important. You can do the evaluation based on thinking about
all the questions.
{metrics}

Start your analysis now.
Finally, in the last line of your output, put \"sat\" if the output satisfies all the constraints and \"unsat\" otherwise.
\end{lstlisting}

\end{baselineprompt}

\begin{travelagent}{Travel Agent Prompt}
    \begin{lstlisting}[breaklines=true,breakautoindent=false]
Please plan me a trip from New York City, NY to San Francisco, CA.
You must visit 3 US cities in between.
For each city, you must cover one of their tourist attractions and cover one dish.
The dish must be vegetarian.
Please output in this format, for each city:
    ```
    city:
    tourist_attraction:
    dish:
    ```
Please also include the start and end cities in your response.
\end{lstlisting}

\end{travelagent}
\clearpage
\twocolumn

\onecolumn
\section{Examples from Advanced LLM-as-a-Verifier Baseline}
\label{sec:llm-baseline-example}

\begin{gepaexample}{A thinking trace from \gepabaseline{}}
  \begin{lstlisting}[breaklines=true,breakautoindent=false]
Instruction:
Please write a piece of text considering all these constraints:
Please write in this topic: the benefits of agile project management
Please write in this tone: informal
Include these keywords, check for string inclusion regardless of capitalization: scrum,kanban,iterative delivery,sprint velocity,stakeholder feedback,backlog prioritization,time-to-market,continuous improvement,cross-functional teams,adaptability
Please consider this word count constraint: around 530 words (within 10 words difference is ok), this does not apply to title and subsection title lines, which are defined as: lines that start with at least one #.

Output:
# Why Agile Project Management Is Basically Your Team's Secret Superpower

## Why agile slaps (lol)
Real talk: agile project management is like switching your team from flip-flops to rocket shoes. Instead of slogging through giant plans, you move with adaptability, keep work visible, and learn while shipping. The secret sauce is iterative delivery: carve off a thin slice, build it, show it, tweak it, repeat. Every iteration pulls you closer to what users actually care about. Because you're testing ideas in the wild, you waste less time, you dodge big-bang failures, and momentum snowballs instead of stalling out.

## Faster value, fewer facepalms
That loop crushes time-to-market, because you're shipping the next most valuable slice, not arguing about everything at once. Backlog prioritization is basically your reality filter: line up outcomes, kill the fluff, and put risky, high-impact bets near the front. Then you push them through fast and ask for blunt stakeholder feedback, early and often. The cool part is how forgiving it feels; when you learn something new, you pivot without nuking the whole plan. Progress stays visible, and confidence climbs with every tiny win.

## Scrum rhythm
If your crew vibes with rhythm, scrum gives you a tidy heartbeat. You plan a sprint, slice stories thin, estimate just enough, and track sprint velocity so you don't commit like heroes and deliver like zombies. Daily syncs keep blockers exposed, and a short demo shows the work while it's still malleable. Retros aren't therapy; they're a factory for tweaks that boost flow, quality, and morale. Over a few cycles, predictability improves, surprises get smaller, and trust gets very real, very fast.

## Kanban flow
If your world changes hourly, kanban is buttery-smooth flow. You visualize the stream of work, cap WIP so multitasking doesn't fry everyone, and chase shorter cycle times instead of calendar-based sprints. It's still iterative delivery; you just pull the next highest-value item and keep the conveyor moving. When bottlenecks show up, the board tattles instantly, and you rebalance before chaos spreads. It's perfect for support-heavy teams, ops, or mixed pipelines where adaptability matters more than ceremony, and steady throughput beats big, thumpy releases.

## Teams that actually collaborate
The real magic happens with cross-functional teams. Designers, engineers, QA, analysts, and product folks crowd around the same goals, kill handoffs, and solve problems together. Decisions move faster when the right brains are in the room, and quality jumps because learning flows both ways. Pairing, shared ownership, and frequent demos shrink misunderstandings before they turn into rework. And when something breaks, nobody plays hot potato; the team swarms, fixes it, and updates the playbook so the next lap is cleaner and calmer.

## Metrics without the migraine
Agile isn't cowboy chaos; it's disciplined learning. You measure what matters-lead time, defects, sprint velocity, cycle time-and use continuous improvement to shave friction like a pro. Small experiments, tight feedback loops, and honest retros turn "oops" into upgraded routines. Keep backlog prioritization ruthless, keep stakeholder feedback unfiltered, and keep goals outcome-shaped, not output-shaped. Do that and the roadmap stops being a wild guess. Your time-to-market drops, quality steadies, and the team's confidence grows because the system rewards curiosity, courage, and steady, thoughtful shipping.

## Final vibes
Bottom line: pick scrum when cadence helps you forecast, pick kanban when flow rules your world, and don't be shy about blending both. Keep the feedback coming, keep learning visible, and let value, not vanity metrics, drive the show.

GEPA Chain-of-Thought Traces:

Let's systematically apply the evaluation process step by step.

**Step 1: Identify All Explicit Constraints from the Instruction**

1. **Topic Requirement:** The benefits of agile project management.
2. **Tone Requirement:** Informal.
3. **Keyword Inclusion:** Must include exactly these strings (case-insensitive): scrum, kanban, iterative delivery, sprint velocity, stakeholder feedback, backlog prioritization, time-to-market, continuous improvement, cross-functional teams, adaptability.
4. **Word Count Constraint:** Around 530 words for the body (all lines except those starting with at least one #), allowable range: 520-540 words.
5. **Title/Subsection Definition:** Lines that start with at least one # are considered titles/subsection titles and are excluded from word count.

---

**Step 2: Examine the Output Against Each Constraint**

1. **Topic Requirement: "the benefits of agile project management"**
   - The piece obviously addresses agile project management and its benefits (improved time-to-market, adaptability, collaboration, visibility, feedback, faster delivery, etc.), with various aspects covered in separate subsections.  
   - **Met.**

2. **Tone Requirement: Informal**
   - The piece employs an informal tone throughout (e.g., "real talk", "lol", "vibes", "slaps", "heroes and deliver like zombies", "cowboy chaos", "wild guess", casual phrasing and idioms, etc.).  
   - **Met.**

3. **Keyword Inclusion:** Check if all specified keywords appear (case-insensitive string match).
   - *scrum* - appears in "pick scrum", "scrum gives", etc.
   - *kanban* - appears in "pick kanban", "kanban is buttery-smooth", etc.
   - *iterative delivery* - appears in "The secret sauce is iterative delivery" and "It's still iterative delivery"
   - *sprint velocity* - appears in "track sprint velocity", "sprint velocity, cycle time"
   - *stakeholder feedback* - appears in "stakeholder feedback, early and often", "keep stakeholder feedback unfiltered"
   - *backlog prioritization* - appears in "Backlog prioritization is basically your reality filter", "Keep backlog prioritization ruthless"
   - *time-to-market* - appears in "crushes time-to-market", "Your time-to-market drops"
   - *continuous improvement* - appears in "use continuous improvement to shave friction"
   - *cross-functional teams* - appears in "The real magic happens with cross-functional teams."
   - *adaptability* - appears in "You move with adaptability", "where adaptability matters"
   - **All keywords are present. Met.**

4. **Word Count Constraint:** Around 530 words, within a range of 520-540 for the body only; title and subsection lines (lines starting with #) are excluded.
   - Let's count the words in body lines:
     - The output has several title/subsection lines ("#", "##"), those don't count.
     - For each non-# line, count the words.

     I'll count section by section and sum:

     - Section 1 (after first # and ##):
       - "Real talk: agile project management is like switching your team from flip-flops to rocket shoes. Instead of slogging through giant plans, you move with adaptability, keep work visible, and learn while shipping. The secret sauce is iterative delivery: carve off a thin slice, build it, show it, tweak it, repeat. Every iteration pulls you closer to what users actually care about. Because you're testing ideas in the wild, you waste less time, you dodge big-bang failures, and momentum snowballs instead of stalling out."
         - 1st: 17
         - 2nd: 25
         - 3rd: 26
         - 4th: 14
         - 5th: 30
         - **Total:** 17+25+26+14+30 = 112

     - Section 2:
       - "That loop crushes time-to-market, because you're shipping the next most valuable slice, not arguing about everything at once. Backlog prioritization is basically your reality filter: line up outcomes, kill the fluff, and put risky, high-impact bets near the front. Then you push them through fast and ask for blunt stakeholder feedback, early and often. The cool part is how forgiving it feels; when you learn something new, you pivot without nuking the whole plan. Progress stays visible, and confidence climbs with every tiny win."
         - 1st: 26
         - 2nd: 30
         - 3rd: 21
         - 4th: 29
         - 5th: 13
         - **Total:** 26+30+21+29+13 = 119

     - Section 3:
       - "If your crew vibes with rhythm, scrum gives you a tidy heartbeat. You plan a sprint, slice stories thin, estimate just enough, and track sprint velocity so you don't commit like heroes and deliver like zombies. Daily syncs keep blockers exposed, and a short demo shows the work while it's still malleable. Retros aren't therapy; they're a factory for tweaks that boost flow, quality, and morale. Over a few cycles, predictability improves, surprises get smaller, and trust gets very real, very fast."
         - 1st: 13
         - 2nd: 29
         - 3rd: 22
         - 4th: 23
         - 5th: 23
         - **Total:** 13+29+22+23+23 = 110

     - Section 4:
       - "If your world changes hourly, kanban is buttery-smooth flow. You visualize the stream of work, cap WIP so multitasking doesn't fry everyone, and chase shorter cycle times instead of calendar-based sprints. It's still iterative delivery; you just pull the next highest-value item and keep the conveyor moving. When bottlenecks show up, the board tattles instantly, and you rebalance before chaos spreads. It's perfect for support-heavy teams, ops, or mixed pipelines where adaptability matters more than ceremony, and steady throughput beats big, thumpy releases."
         - 1st: 10
         - 2nd: 28
         - 3rd: 24
         - 4th: 22
         - 5th: 33
         - **Total:** 10+28+24+22+33 = 117

     - Section 5:
       - "The real magic happens with cross-functional teams. Designers, engineers, QA, analysts, and product folks crowd around the same goals, kill handoffs, and solve problems together. Decisions move faster when the right brains are in the room, and quality jumps because learning flows both ways. Pairing, shared ownership, and frequent demos shrink misunderstandings before they turn into rework. And when something breaks, nobody plays hot potato; the team swarms, fixes it, and updates the playbook so the next lap is cleaner and calmer."
         - 1st: 8
         - 2nd: 22
         - 3rd: 22
         - 4th: 18
         - 5th: 30
         - **Total:** 8+22+22+18+30 = 100

     - Section 6:
       - "Agile isn't cowboy chaos; it's disciplined learning. You measure what matters-lead time, defects, sprint velocity, cycle time-and use continuous improvement to shave friction like a pro. Small experiments, tight feedback loops, and honest retros turn "oops" into upgraded routines. Keep backlog prioritization ruthless, keep stakeholder feedback unfiltered, and keep goals outcome-shaped, not output-shaped. Do that and the roadmap stops being a wild guess. Your time-to-market drops, quality steadies, and the team's confidence grows because the system rewards curiosity, courage, and steady, thoughtful shipping."
         - 1st: 8
         - 2nd: 21
         - 3rd: 18
         - 4th: 21
         - 5th: 13
         - 6th: 28
         - **Total:** 8+21+18+21+13+28 = 109

     - Section 7 (Final Vibes):
       - "Bottom line: pick scrum when cadence helps you forecast, pick kanban when flow rules your world, and don't be shy about blending both. Keep the feedback coming, keep learning visible, and let value, not vanity metrics, drive the show."
         - 1st: 31
         - 2nd: 22
         - **Total:** 31+22 = 53

     **Sum all sections:** 112+119+110+117+100+109+53 = 720

     There seems to be an error because the total is much higher than 530. Let's double-check with a more accurate word count (I will paste all body lines without #, count the words):

     Real talk: agile project management is like switching your team from flip-flops to rocket shoes.
     Instead of slogging through giant plans, you move with adaptability, keep work visible, and learn while shipping.
     The secret sauce is iterative delivery: carve off a thin slice, build it, show it, tweak it, repeat.
     Every iteration pulls you closer to what users actually care about.
     Because you're testing ideas in the wild, you waste less time, you dodge big-bang failures, and momentum snowballs instead of stalling out.

     That loop crushes time-to-market, because you're shipping the next most valuable slice, not arguing about everything at once.
     Backlog prioritization is basically your reality filter: line up outcomes, kill the fluff, and put risky, high-impact bets near the front.
     Then you push them through fast and ask for blunt stakeholder feedback, early and often.
     The cool part is how forgiving it feels; when you learn something new, you pivot without nuking the whole plan.
     Progress stays visible, and confidence climbs with every tiny win.

     If your crew vibes with rhythm, scrum gives you a tidy heartbeat.
     You plan a sprint, slice stories thin, estimate just enough, and track sprint velocity so you don't commit like heroes and deliver like zombies.
     Daily syncs keep blockers exposed, and a short demo shows the work while it's still malleable.
     Retros aren't therapy; they're a factory for tweaks that boost flow, quality, and morale.
     Over a few cycles, predictability improves, surprises get smaller, and trust gets very real, very fast.

     If your world changes hourly, kanban is buttery-smooth flow.
     You visualize the stream of work, cap WIP so multitasking doesn't fry everyone, and chase shorter cycle times instead of calendar-based sprints.
     It's still iterative delivery; you just pull the next highest-value item and keep the conveyor moving.
     When bottlenecks show up, the board tattles instantly, and you rebalance before chaos spreads.
     It's perfect for support-heavy teams, ops, or mixed pipelines where adaptability matters more than ceremony, and steady throughput beats big, thumpy releases.

     The real magic happens with cross-functional teams.
     Designers, engineers, QA, analysts, and product folks crowd around the same goals, kill handoffs, and solve problems together.
     Decisions move faster when the right brains are in the room, and quality jumps because learning flows both ways.
     Pairing, shared ownership, and frequent demos shrink misunderstandings before they turn into rework.
     And when something breaks, nobody plays hot potato; the team swarms, fixes it, and updates the playbook so the next lap is cleaner and calmer.

     Agile isn't cowboy chaos; it's disciplined learning.
     You measure what matters-lead time, defects, sprint velocity, cycle time-and use continuous improvement to shave friction like a pro.
     Small experiments, tight feedback loops, and honest retros turn "oops" into upgraded routines.
     Keep backlog prioritization ruthless, keep stakeholder feedback unfiltered, and keep goals outcome-shaped, not output-shaped.
     Do that and the roadmap stops being a wild guess.
     Your time-to-market drops, quality steadies, and the team's confidence grows because the system rewards curiosity, courage, and steady, thoughtful shipping.

     Bottom line: pick scrum when cadence helps you forecast, pick kanban when flow rules your world, and don't be shy about blending both.
     Keep the feedback coming, keep learning visible, and let value, not vanity metrics, drive the show.

Total number of body lines: 37.

Now, count all words in each line for the full count:

1. Real talk: agile project management is like switching your team from flip-flops to rocket shoes. 20
2. Instead of slogging through giant plans, you move with adaptability, keep work visible, and learn while shipping. 23
3. The secret sauce is iterative delivery: carve off a thin slice, build it, show it, tweak it, repeat. 21
4. Every iteration pulls you closer to what users actually care about. 12
5. Because you're testing ideas in the wild, you waste less time, you dodge big-bang failures, and momentum snowballs instead of stalling out. 24

6. That loop crushes time-to-market, because you're shipping the next most valuable slice, not arguing about everything at once. 24
7. Backlog prioritization is basically your reality filter: line up outcomes, kill the fluff, and put risky, high-impact bets near the front. 27
8. Then you push them through fast and ask for blunt stakeholder feedback, early and often. 17
9. The cool part is how forgiving it feels; when you learn something new, you pivot without nuking the whole plan. 24
10. Progress stays visible, and confidence climbs with every tiny win. 11

11. If your crew vibes with rhythm, scrum gives you a tidy heartbeat. 14
12. You plan a sprint, slice stories thin, estimate just enough, and track sprint velocity so you don't commit like heroes and deliver like zombies. 29
13. Daily syncs keep blockers exposed, and a short demo shows the work while it's still malleable. 20
14. Retros aren't therapy; they're a factory for tweaks that boost flow, quality, and morale. 18
15. Over a few cycles, predictability improves, surprises get smaller, and trust gets very real, very fast. 22

16. If your world changes hourly, kanban is buttery-smooth flow. 10
17. You visualize the stream of work, cap WIP so multitasking doesn't fry everyone, and chase shorter cycle times instead of calendar-based sprints. 26
18. It's still iterative delivery; you just pull the next highest-value item and keep the conveyor moving. 21
19. When bottlenecks show up, the board tattles instantly, and you rebalance before chaos spreads. 19
20. It's perfect for support-heavy teams, ops, or mixed pipelines where adaptability matters more than ceremony, and steady throughput beats big, thumpy releases. 28

21. The real magic happens with cross-functional teams. 8
22. Designers, engineers, QA, analysts, and product folks crowd around the same goals, kill handoffs, and solve problems together. 23
23. Decisions move faster when the right brains are in the room, and quality jumps because learning flows both ways. 23
24. Pairing, shared ownership, and frequent demos shrink misunderstandings before they turn into rework. 15
25. And when something breaks, nobody plays hot potato; the team swarms, fixes it, and updates the playbook so the next lap is cleaner and calmer. 27

26. Agile isn't cowboy chaos; it's disciplined learning. 8
27. You measure what matters-lead time, defects, sprint velocity, cycle time-and use continuous improvement to shave friction like a pro. 22
28. Small experiments, tight feedback loops, and honest retros turn "oops" into upgraded routines. 16
29. Keep backlog prioritization ruthless, keep stakeholder feedback unfiltered, and keep goals outcome-shaped, not output-shaped. 19
30. Do that and the roadmap stops being a wild guess. 11
31. Your time-to-market drops, quality steadies, and the team's confidence grows because the system rewards curiosity, courage, and steady, thoughtful shipping. 24

32. Bottom line: pick scrum when cadence helps you forecast, pick kanban when flow rules your world, and don't be shy about blending both. 29
33. Keep the feedback coming, keep learning visible, and let value, not vanity metrics, drive the show. 18

Now add up all these counts:

20+23+21+12+24 = 100
24+27+17+24+11 = 103
14+29+20+18+22 = 103
10+26+21+19+28 = 104
8+23+23+15+27 = 96
8+22+16+19+11+24 = 100
29+18 = 47

Total: 100+103+103+104+96+100+47 = **653**

That means the main body is 653 words, which is **well above the allowed range of 520-540 words**.

---

**Step 3: Final Decision**

- **Constraint 1 (Topic):** Met.
- **Constraint 2 (Tone):** Met.
- **Constraint 3 (Keywords):** Met.
- **Constraint 4 (Word Count in Body Only):** **Not met.** (Body is 653 words, allowed is 520-540.)

Since the word count constraint is violated, the output **does not fully satisfy all explicit constraints** specified in the instruction.

**Final Verdict:**  
False  
unsat
\end{lstlisting}

\end{gepaexample}

\begin{nsvifsolveronlyexample}{A failure case from \toolnamesolver{} that misses a word count constraint}
  \begin{lstlisting}[breaklines=true,breakautoindent=false]
Instruction:
Please write a piece of text considering all these constraints:
Please write in this topic: the benefits of agile project management
Please write in this tone: pessimistic
Include these keywords, check for string inclusion regardless of capitalization: scrum,kanban,iterative delivery,sprint velocity,stakeholder feedback,backlog prioritization,time-to-market,continuous improvement,cross-functional teams,adaptability
Exclude these keywords, check for string exclusion regardless of capitalization: waterfall methodology,gantt charts,pert,critical path method,stage-gate,prince2,heavy documentation,iso 9001,rigid scope,cmmi
Include this as the title of the text, titles are lines that start with only one #: Unlocking Success: The Benefits of Agile Project Management
Include these subsection titles in the text, subsection titles are lines start with more than one #: Introduction to Agile Principles,Key Advantages Over Traditional Methodologies,Enhancing Team Collaboration and Communication,Faster Delivery and Adaptability,Continuous Improvement Through Iterations,Stakeholder Engagement and Feedback,Case Studies of Agile Success

Output:
# Unlocking Success: The Benefits of Agile Project Management

## Introduction to Agile Principles
If projects are destined to wobble, at least agile admits it. By embracing scrum and kanban, teams resign themselves to a cycle of small gambles rather than one grand catastrophe. Through iterative delivery, work lurches forward in bite-sized increments, never as fast as anyone hopes, yet rarely grinding to a total halt. cross-functional teams stumble toward shared understanding, trading neat illusions for messy reality. With adaptability as the lone lifeline, agile tries to contain chaos rather than conquer it, accepting that clarity will always arrive late and incomplete.

## Key Advantages Over Traditional Methodologies
Compared to rigid, plan-first dogma that promises certainty and delivers shock, agile offers narrower promises and slightly fewer disappointments:
- backlog prioritization drags attention toward what's burning now, not what looked impressive months ago.
- sprint velocity is a fragile yardstick, frequently misread, yet it still beats flying blind.
- stakeholder feedback arrives early, which mostly means you learn what's wrong sooner, giving you the privilege of fixing it before morale evaporates.
- The net effect: a marginally better time-to-market, which is to say the product limps out the door before the market slams it shut.

## Enhancing Team Collaboration and Communication
It's astounding how many misunderstandings can flourish in a single week. Agile tries to prune them. cross-functional teams stare down the same backlog and argue in smaller doses, so misalignment festers less. The conversations are repetitive, sometimes tedious, and yet they expose flawed assumptions before those flaws calcify into disasters. Communication doesn't become inspiring; it merely becomes slightly less harmful, which, sadly, is progress.

## Faster Delivery and Adaptability
Speed in agile isn't glamorous; it's a series of reluctant compromises. You trade grand plans for incremental releases that inch toward usefulness. time-to-market improves not because the path is smooth, but because the detours are shorter. sprint velocity offers a dim flashlight: weak, flickering, but better than the dark. adaptability ensures that when priorities whiplash-as they always do-you bend instead of snapping, and the product survives another week.

## Continuous Improvement Through Iterations
continuous improvement sounds optimistic, but in practice it's more like slowly tightening loose bolts on a ship that keeps taking on water. iterative delivery gives you frequent chances to correct course, which mostly means discovering new problems while you patch the old ones. Retrospectives unearth the same root causes with grim regularity, and yet each cycle shaves off a little waste, a little confusion, a little pain. It's not heroism; it's maintenance of the barely tolerable.

## Stakeholder Engagement and Feedback
stakeholder feedback arrives relentlessly-early, often, and rarely aligned. This is both the burden and the mercy of agile. backlog prioritization channels the noise into a queue where the loudest demands face off with the most urgent risks. The result is imperfect triage: the product grows in the directions it must, not always in the directions anyone loves. Still, each iteration invites correction before the floor collapses.

## Case Studies of Agile Success
- A mid-sized fintech team adopted scrum, shaved their time-to-market by a quarter, and still felt behind. The "success" was simple: customers got something usable before they lost patience, which is a minor miracle.
- A healthcare startup shifted to kanban, exposing bottlenecks that had long been ignored. Nothing became easy, but blockers surfaced sooner and died quicker, shrinking delays from months to weeks-hardly thrilling, but decisively less painful.
- An enterprise platform group tracked sprint velocity and tied it to modest capacity planning. Estimates remained shaky, yet missed expectations became smaller misses. By enduring continual course corrections, they delivered features that barely met demand-just in time to matter.

Z3 program by NSVIF-Neural:
from z3 import *

solver = Solver()

# Boolean variables for each constraint
is_topic_benefits_of_agile_project_management = Bool('is_topic_benefits_of_agile_project_management')
is_tone_pessimistic = Bool('is_tone_pessimistic')
includes_keywords = Bool('includes_keywords')
excludes_keywords = Bool('excludes_keywords')
includes_title = Bool('includes_title')
includes_subsection_titles = Bool('includes_subsection_titles')

# Initialize all constraints to True
solver.add(is_topic_benefits_of_agile_project_management == True)
solver.add(is_tone_pessimistic == True)
solver.add(includes_keywords == True)
solver.add(excludes_keywords == True)
solver.add(includes_title == True)
solver.add(includes_subsection_titles == True)

# Actual results from verifier module
# 1. Topic
solver.add(is_topic_benefits_of_agile_project_management == True) # Text discusses benefits of agile project management
# 2. Tone
solver.add(is_tone_pessimistic == True) # Text is written in a pessimistic tone
# 3. Includes required keywords (case-insensitive)
required_keywords = ["scrum","kanban","iterative delivery","sprint velocity","stakeholder feedback","backlog prioritization","time-to-market","continuous improvement","cross-functional teams","adaptability"]
lower_text = '''# Unlocking Success: The Benefits of Agile Project Management

## Introduction to Agile Principles
If projects are destined to wobble, at least agile admits it. By embracing scrum and kanban, teams resign themselves to a cycle of small gambles rather than one grand catastrophe. Through iterative delivery, work lurches forward in bite-sized increments, never as fast as anyone hopes, yet rarely grinding to a total halt. cross-functional teams stumble toward shared understanding, trading neat illusions for messy reality. With adaptability as the lone lifeline, agile tries to contain chaos rather than conquer it, accepting that clarity will always arrive late and incomplete.

## Key Advantages Over Traditional Methodologies
Compared to rigid, plan-first dogma that promises certainty and delivers shock, agile offers narrower promises and slightly fewer disappointments:
- backlog prioritization drags attention toward what's burning now, not what looked impressive months ago.
- sprint velocity is a fragile yardstick, frequently misread, yet it still beats flying blind.
- stakeholder feedback arrives early, which mostly means you learn what's wrong sooner, giving you the privilege of fixing it before morale evaporates.
- The net effect: a marginally better time-to-market, which is to say the product limps out the door before the market slams it shut.

## Enhancing Team Collaboration and Communication
It's astounding how many misunderstandings can flourish in a single week. Agile tries to prune them. cross-functional teams stare down the same backlog and argue in smaller doses, so misalignment festers less. The conversations are repetitive, sometimes tedious, and yet they expose flawed assumptions before those flaws calcify into disasters. Communication doesn't become inspiring; it merely becomes slightly less harmful, which, sadly, is progress.

## Faster Delivery and Adaptability
Speed in agile isn't glamorous; it's a series of reluctant compromises. You trade grand plans for incremental releases that inch toward usefulness. time-to-market improves not because the path is smooth, but because the detours are shorter. sprint velocity offers a dim flashlight: weak, flickering, but better than the dark. adaptability ensures that when priorities whiplash-as they always do-you bend instead of snapping, and the product survives another week.

## Continuous Improvement Through Iterations
continuous improvement sounds optimistic, but in practice it's more like slowly tightening loose bolts on a ship that keeps taking on water. iterative delivery gives you frequent chances to correct course, which mostly means discovering new problems while you patch the old ones. Retrospectives unearth the same root causes with grim regularity, and yet each cycle shaves off a little waste, a little confusion, a little pain. It's not heroism; it's maintenance of the barely tolerable.

## Stakeholder Engagement and Feedback
stakeholder feedback arrives relentlessly-early, often, and rarely aligned. This is both the burden and the mercy of agile. backlog prioritization channels the noise into a queue where the loudest demands face off with the most urgent risks. The result is imperfect triage: the product grows in the directions it must, not always in the directions anyone loves. Still, each iteration invites correction before the floor collapses.

## Case Studies of Agile Success
- A mid-sized fintech team adopted scrum, shaved their time-to-market by a quarter, and still felt behind. The "success" was simple: customers got something usable before they lost patience, which is a minor miracle.
- A healthcare startup shifted to kanban, exposing bottlenecks that had long been ignored. Nothing became easy, but blockers surfaced sooner and died quicker, shrinking delays from months to weeks-hardly thrilling, but decisively less painful.
- An enterprise platform group tracked sprint velocity and tied it to modest capacity planning. Estimates remained shaky, yet missed expectations became smaller misses. By enduring continual course corrections, they delivered features that barely met demand-just in time to matter.'''.lower()
if all(k.lower() in lower_text for k in required_keywords):
    solver.add(includes_keywords == True)
else:
    solver.add(includes_keywords == False)
# 4. Excludes forbidden keywords (case-insensitive)
forbidden_keywords = ["waterfall methodology","gantt charts","pert","critical path method","stage-gate","prince2","heavy documentation","iso 9001","rigid scope","cmmi"]
if all(fk.lower() not in lower_text for fk in forbidden_keywords):
    solver.add(excludes_keywords == True)
else:
    solver.add(excludes_keywords == False)
# 5. Includes required title
required_title = "# Unlocking Success: The Benefits of Agile Project Management"
lines = lower_text.split('\n')
if any(line.strip() == required_title.lower() for line in lines if line.strip().startswith('#') and not line.strip().startswith('##')):
    solver.add(includes_title == True)
else:
    solver.add(includes_title == False)
# 6. Includes all subsection titles
required_subsection_titles = ["## Introduction to Agile Principles","## Key Advantages Over Traditional Methodologies","## Enhancing Team Collaboration and Communication","## Faster Delivery and Adaptability","## Continuous Improvement Through Iterations","## Stakeholder Engagement and Feedback","## Case Studies of Agile Success"]
if all(any(line.strip() == rst.lower() for line in lines if line.strip().startswith('##')) for rst in required_subsection_titles):
    solver.add(includes_subsection_titles == True)
else:
    solver.add(includes_subsection_titles == False)

# Global constraint: all must be satisfied
solver.add(And(is_topic_benefits_of_agile_project_management, is_tone_pessimistic, includes_keywords, excludes_keywords, includes_title, includes_subsection_titles))

if solver.check() == sat:
    print("sat")
else:
    print("unsat")
\end{lstlisting}

\end{nsvifsolveronlyexample}
\clearpage
\twocolumn

\end{document}